# Denoising instrumented mouthguard measurements of head impact kinematics with a convolutional neural network

Xianghao Zhan (1), Yuzhe Liu (1), Nicholas J. Cecchi (1), Ashlyn A. Callan (1), Enora Le Flao (1), Olivier Gevaert (2), Michael M. Zeineh (3), Gerald A. Grant (4), David B. Camarillo (1)

(1) Department of Bioengineering, Stanford University, CA, 94305, USA
(2) Department of Biomedical Data Science, Stanford University, CA, 94305, USA
(3) Department of Radiology, Stanford University, CA, 94305, USA
(4) Department of Neurosurgery, Duke University, NC, 27710, USA

Corresponding author: Yuzhe Liu (yuzheliu@stanford.edu)

## Abstract

Wearable sensors are developed to measure head kinematics but are intrinsically noisy because of the imperfect interface with the human body. Instrumented mouthguards are designed to fit rigidly to upper dentition to measure head kinematics during impacts in studies on traumatic brain injury (TBI). However, deviations of the mouthguard measurement from reference kinematics in laboratory experiments can still be observed due to potential loosening. In this study, we apply deep learning to compensate for the imperfect interfacing and improve the kinematics measurement accuracy with 163 laboratory head impacts partitioned into training, validation, and test sets. On the training set, we developed a set of one-dimensional convolutional neural network (1D-CNN) models to denoise the mouthguard kinematics measurements along three spatial axes of linear acceleration and angular velocity. The performance was evaluated on three levels. On the kinematics level, the denoised kinematics are better correlated with gold-standard reference kinematics with significantly-reduced pointwise root mean squared error (mean error reduction: 36%) and peak absolute error (mean error reduction: 56%). On four out of six brain injury criteria calculated from the kinematics, denoising led to significantly reduced absolute errors (mean error reduction: 82%). On the level of tissue strain and strain rate calculated via finite element modeling, denoising led to significantly reduced absolute error in maximum principal strain and maximum principal strain rate (mean error reduction: 35% and 69%, respectively). The lesser denoising effect at the strain/strain rate level when compared with that at the kinematics-level reflects the brain's filtering effect on head accelerations. On an on-field dataset of 118 college football impacts, we performed a blind test of the 1D-CNN denoising models, which showed reductions in strain and strain rate after denoising. In an exploratory study on 413 post-mortem human subject impacts, where the lower dentition directly impacted the mouthguard, similar denoising effects were observed and the peak kinematics after denoising were more accurate (relative error reduction for top 10% noisiest impacts was 75.6%). This study provides the models to denoise mouthguard kinematics to improve detection of head impacts and TBI risk evaluation, which can be extended to denoise other sensors measuring kinematics.

**Keywords:** traumatic brain injury, head kinematics, deep learning, signal denoising

## Introduction

Traumatic brain injury (TBI) is a global health threat. Humans are exposed to TBI risks from situations such as falling, transportation accidents, occupational hazards, military combat, sports, and recreation [1]. In 2016 alone, over 27 million cases of TBI were reported globally, 80% of which were characterized as "mild" (mTBI), which means the immediate symptoms, including loss of consciousness, headache or fatigue, may recover in the short term. However, patients often suffer from longer-term pathophysiological changes of the brain [1,2]. In the clinical TBI classification, patients with a Glasgow Coma Scale score of larger than 12 are classified as mTBI sufferers and concussion is generally recognized as the clinically diagnosed mTBI [3]. Long after the acute trauma, single or repeated mTBIs are primary contributors in 3-15% of all dementia cases including Alzheimer's disease, Parkinson's disease, and chronic traumatic encephalopathy [4,5]. The diffuse traumatic injury caused by the inertial movement of the brain after the head sustains rapid acceleration and deceleration is the dominant cause of mTBI in humans [6]. This calls for the development of accurate and immediate measurement of head movement for the early detection of mTBI and intervention. The fast diagnosis and early warning of mTBI [7] are significant to helping prevent repetitive mTBI, because the real-time intervention after early detection can significantly improve the outcome [8]. For example, on the sports field [9], once a dangerous impact is detected, a football player could take a rest before further brain damage would accumulate from future head impacts in continued play.

To estimate the TBI risks, brain injury criteria (BIC) have been developed and used [10]. Examples are the head injury criterion (HIC) developed by Versace et al. [11] and brain injury criterion (BrIC) developed by Takhounts et al. [12]. Physiologically, the brain is damaged because of the inertial movement of the brain after the rapid head movement causes tissue-level deformation. Many recent studies have shown that brain strain (maximum principal strain, MPS) and strain rate (MPS rate, MPSR) based on brain-physics-based finite element modeling (FEM) are promising mechanical parameters that correlate with mTBI pathologies including blood-brain-barrier disruption and traumatic axonal injury [13,14].

The head impact kinematics are the input to compute the various BIC, brain strain and strain rate. To calculate these injury metrics, researchers have developed different wearable sensor technologies to measure head kinematics with systems like the Head Impact Telemetry System (HITS) [15], Xpatch [16], headband/skullcap mounted sensors [17] and instrumented mouthguards [18-20]. However, wearable sensors are intrinsically noisy because of the imperfect interface between the sensors and the human body. Therefore, unlike the HITS, Xpatch, and headband/skullcap systems that attach sensors to headgear or the skin where the relative motion can interfere with the measurements for the skull kinematics, instrumented mouthguards attach the sensors to the dentition, which enables rigid coupling with the skull, therefore yielding more precise measurement of the head's rotation. Recently, the brain strain calculated based on mouthguard-measured head kinematics was found to correlate well with resulting mTBI [21,22], which suggests the feasibility of using instrumented mouthguards to help diagnose mTBI with brain strain and strain rate.

Although instrumented mouthguards have shown the capability to precisely measure head kinematics [19] that correlate well with the mTBI results when compared with other measurement technologies, the kinematics measurements are still different from the reference kinematics measured by sensors directly installed into an anthropomorphic test device (ATD) headform. This is due to the potential loosening of the mouthguard and electronic noise from the sensors during the impact. Ideally, the instrumented mouthguard should be rigidly attached to the dentition. However, in real-world scenarios, the loosening of the mouthguard and other electronic noise can lead to kinematics measurements that are not fully representative of the real head kinematics [9]. The relative motion of the mouthguard can be regarded as a type of noise that pollutes the measurement of head kinematics. Furthermore, the electronic noise in the measurement system during impacts may also be a source of noise in the kinematics measurements. For example, Liu et al. compared five different types of instrumented mouthguards on laboratory impacts [18]. With the measurement of ATD sensors as the reference, the mouthguards show an average of 2.5% to 32.4% relative error in the peak of magnitude in linear acceleration measurements and an average of 2.3% to 7.6% relative error in the peak of magnitude in angular velocity measurement. The errors in the kinematics can further lead to approximately 10% errors in the estimation of 95th percentile MPS [18]. Recently, the variations of accuracies of head kinematics data have been noticed and the Consensus Head Acceleration Measurement Practices (CHAMP) [50-52] was reached by researchers in this field to improve the quality of head impact dataset in TBI studies. Therefore, it will benefit the whole TBI research community to develop models to denoise the kinematics measurements by compensating the imperfect interfacing of wearable sensors.

Deep learning, which employs diverse structures of artificial neural networks to learn the mapping from the input data to output data through the training data, has recently shown its effectiveness in signal denoising in examples such as electrocardiography signals [23,24]. There have also been multiple applications of deep learning in TBI research: for example, Zhan et al. and Ghazi et al. have developed deep learning head models (DLHMs) to substitute time-consuming FEM to rapidly and accurately estimate the whole-brain strain [25,26]. However, there has been no previous research leveraging deep learning in the denoising of head impact kinematics. With impact datasets, using deep learning to denoise the kinematics measured by instrumented mouthguards provides a viable tool to obtain head kinematics with higher accuracy.

In this study, we hypothesize that deep learning models can learn the patterns of noise from the large quantities of kinematics data. We applied deep learning to denoise the mouthguard kinematics to compensate for the imperfect interfacing between the sensors and wearers and get the kinematics measurements that better correlate with the reference head kinematics measured by sensors in an ATD headform. With 163 independent laboratory impacts at different locations and velocities from two datasets (a standard protocol dataset and a no-repeat dataset) generated with a pneumatic impactor, and the mouthguard/ATD sensor kinematics measurements associated with the head impacts (Video 1), we firstly partitioned the impacts into training (113 impacts before data augmentation, for modeling training), validation (25 impacts for hyperparameter tuning) and test sets (25 impacts for model evaluation). Then, we performed data augmentation with a sliding window (width: 100ms, stride: 5ms) and trained the one-dimensional convolutional neural network (1D-CNN) to denoise the mouthguard kinematics measurements. The model performance was evaluated on a hold-out test set on three levels: the kinematics level (Results Section 1), the BIC level (Results Section 2), and the brain strain and strain rate level (Results Section 3), when these quantities are computed from the kinematics measurements with/without denoising. The results of denoising showed the effectiveness of 1D-CNN in denoising the mouthguard kinematics measurement to get kinematics signals better correlated with ATD sensor kinematics measurement as well as the significantly more accurate BIC and strain (MPS) and strain rate (MPSR) estimation. Additionally, on a dataset of 118 on-field college football impacts, we applied the 1D-CNN mouthguard denoising model to perform a blind test (Results Section 4). Furthermore, we did an exploratory study to test the denoising models' effectiveness under a worst-case scenario: the lower dentition directly impacted the mouthguard in a dataset of 413 post-mortem human subject (PMHS) head impacts (Results Section 5). The results indicate that the BIC and strain/strain rate values calculated on the denoised mouthguard measurements are significantly different from those calculated on the original signals. In PMHS the denoising effectively reduced the error in kinematics and particularly for the impact with higher noisy components. Finally, a comparison between the 1D-CNN denoising models and the signal filtering was performed (Results Section 6). In summary, the 1D-CNN mouthguard denoising model enables more accurate brain injury risk estimation.

## Results

Based on the impacts collected in our laboratory with a linear pneumatic impactor, we developed six 1D-CNN models to denoise the original head kinematics signals measured by the instrumented mouthguards (1D-CNN model input), with the kinematics measured by the ATD sensors as the ground-truth reference (1D-CNN model output). It should be mentioned that the original signals and the reference signals are filtered with an optimized filter before the development of 1D-CNN models (Method Section 1). The six models matched the six degrees of freedom of the instrumented mouthguard (Methods Section 2). In the result sections, we analyzed the model performance on both the entire test set（25 impacts) and a 9-impact test subset. We collected two datasets for this study. In the standard protocol dataset, there are impacts with the repetitive location and velocity since this dataset was originally used for mouthguard validation (the ATD was impacted at a specific location with a specific velocity for more than 1 times) [18]. As a result, after the dataset partitioning, the impacts from the repeated combination of impact location and impact velocity can be randomly partitioned into training/validation/test sets, which may lead to the overestimation of model accuracy. On the contrary, the no-repeat dataset is composed of impacts from unrepeated combinations of impact location and impact velocity. Therefore, we selected the 9-impact test subset from the no-repeat dataset to evaluate the model performance on the impacts with the combinations of impact location and impact velocity completely unseen in the training and validation set. The best model hyperparameters are listed in Table S1 after the hyperparameter tuning on the validation set (25 impacts) and the comparison of model accuracy of different deep learning structures have been shown in Table S2.

1. **Mouthguard denoising enables the head impact kinematics measurement to be better correlated with the reference kinematics measured by ATD headform sensors.**

After applying the 1D-CNN models on the test impacts, we firstly compared the denoised mouthguard kinematics measurements with the ATD sensor kinematics measurement and the original mouthguard measurements. To visualize the effect of denoising, we plotted the results on three example impacts randomly selected from the 9-impact test subset in Fig. 1. The peak Lin. Acc. magnitude and peak Ang. Vel. magnitude of the ATD sensor kinematics (reference), original mouthguard kinematics, denoised mouthguard kinematics are shown in Fig. 2. To show the error distribution, the Bland Altman plots are shown in the Fig. S1 and S2 to visualize the error distribution over the absolute value of the kinematics components. According to the results, the denoised kinematics measurements are closer to the reference ATD sensor kinematics measurements than the original mouthguard kinematics measurements.

To quantify the effect of denoising, we calculated pointwise mean absolute error (MAE), pointwise root mean squared error (RMSE), peak absolute error (PAE), RMSE of the peak, coefficient of determination ($R^2$) of the peak, signal-to-noise ratio (SNR), Pearson correlation coefficient (r) and Spearman correlation coefficient (SCor) between the reference ATD kinematics measurements and the mouthguard kinematics measurements with/without denoising. The mean values of these metrics averaged over the entire set of test impacts are shown in Table 2 (the metrics on the 9-impact test subset are shown in Table 3). After denoising, the kinematics signals better correlate with the reference kinematics signals, with smaller MAE, RMSE and higher $R^2$ and SNR. Additionally, the errors between the peaks of the mouthguard kinematics measurements and the peaks of the reference kinematics measurements are smaller after denoising.

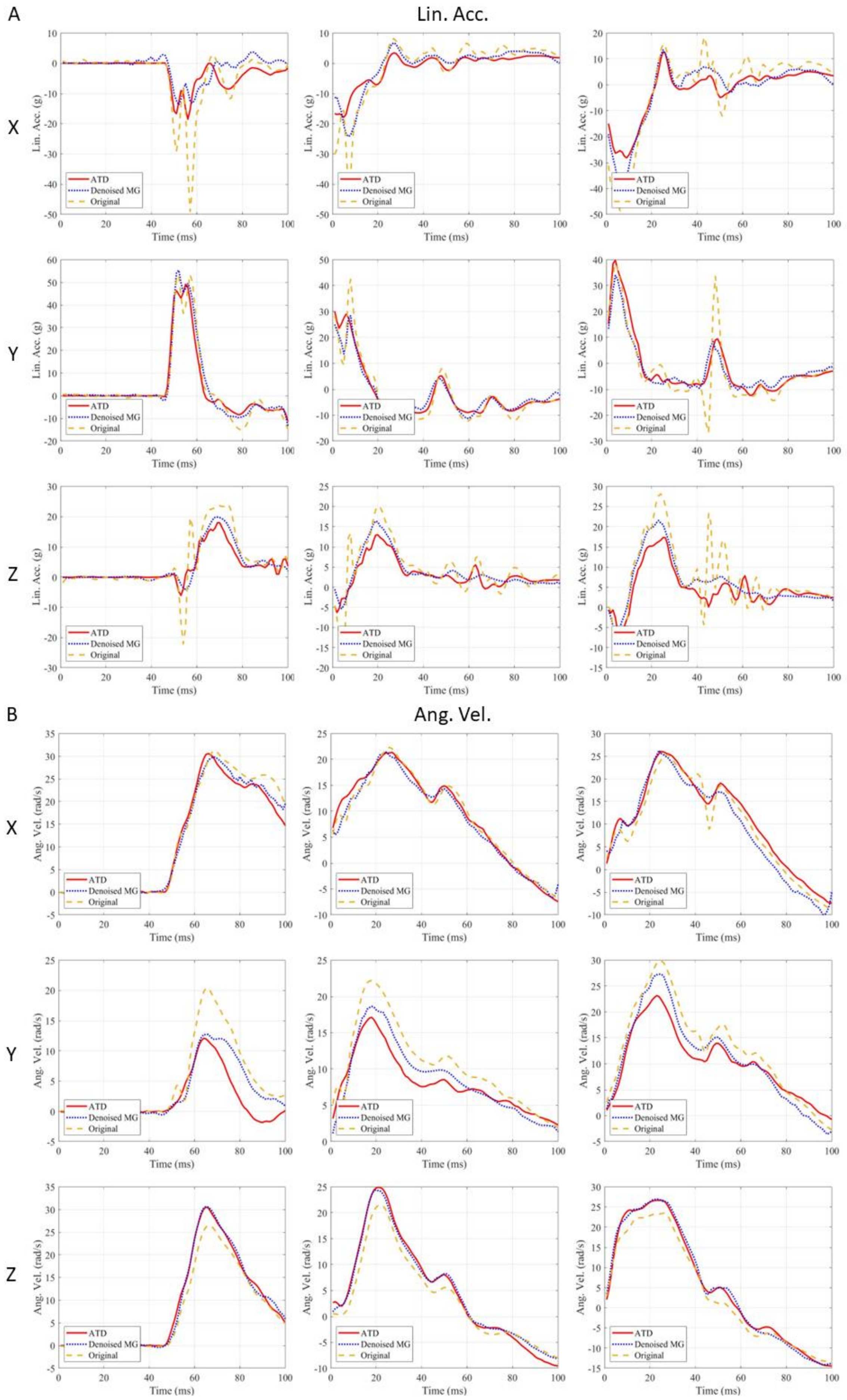

**Figure 1. The denoising effect on the linear acceleration and angular velocity on three example impacts.** The three example impacts are randomly selected from the 9 test impacts originating from the no-repeat dataset. Each row shows the denoising effect on the linear acceleration along one axis (X: posterior-to-anterior, Y: left-to-right, Z: superior-to-inferior). The 9 test impacts from the no-repeat dataset are specifically selected to evaluate the model performance on completely unseen impact directions and impact velocities. ATD: the kinematics measurement from the sensors implanted in the anthropomorphic test device headform; MG: mouthguard measurement.

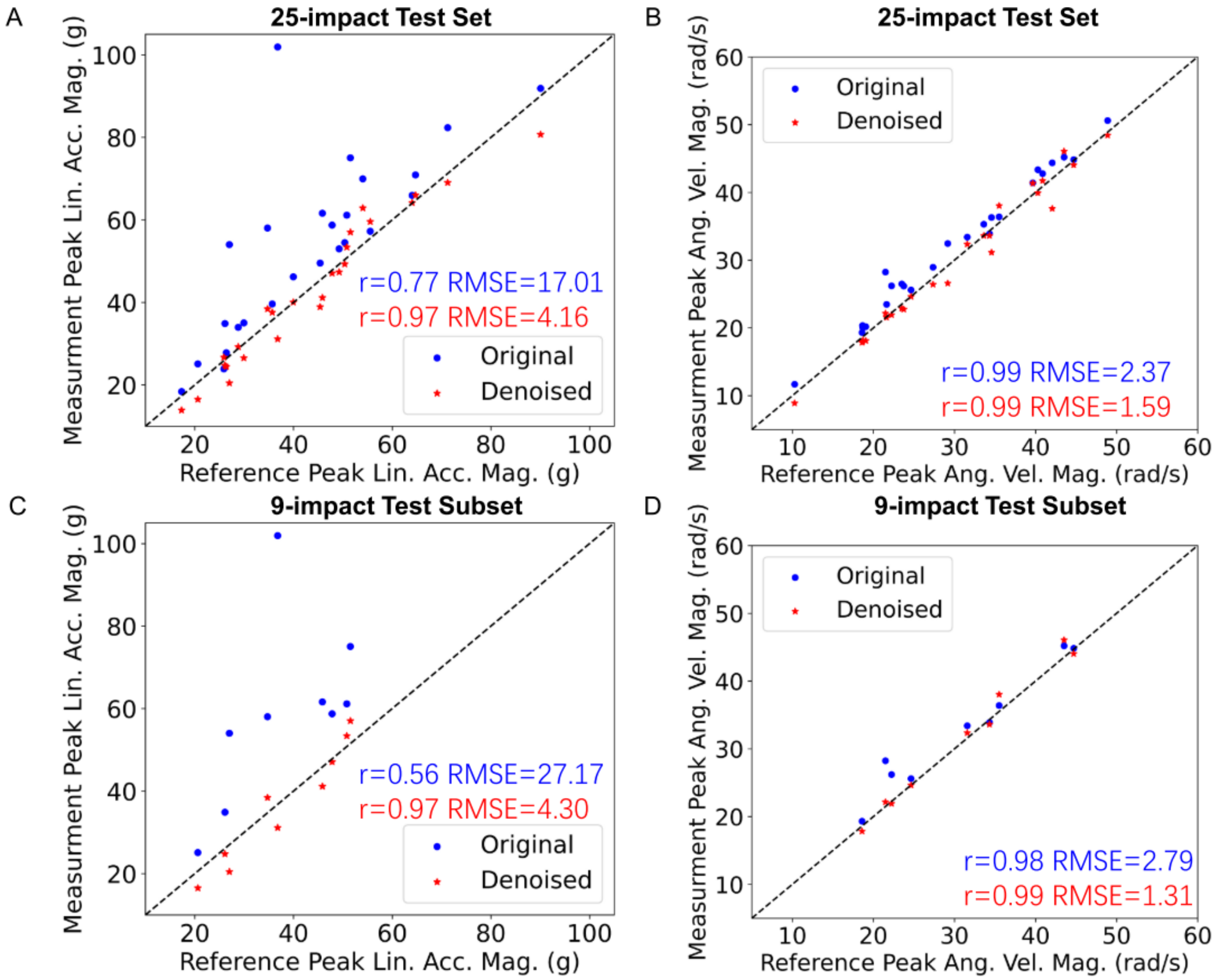

**Figure 2. The peak kinematics magnitude of the reference, original mouthguard measurements, denoised mouthguard measurements.** Results are shown for the peak linear acceleration magnitude (A) and the peak angular velocity magnitude (B) on the entire test set (25 impacts), the peak linear acceleration magnitude (C) and the peak angular velocity magnitude (D) on the 9-impact test subset. The Pearson correlation coefficient (r) and root mean squared error (RMSE) are reported in the figure.

To statistically evaluate the errors between the mouthguard kinematics and the reference kinematics, we plotted the distribution of pointwise RMSE and PAE on the entire test sets (25 impacts) and the test subset (9 test impacts originating from the no-repeat dataset) for 8 kinematics components (X-axis/Y-axis/Z-axis/Magnitude of linear acceleration and angular velocity) in Fig. 3 and Fig. 4. The relative reduction in mean/median errors, as well as the statistical significance, were marked in the figures. On the entire test set, the RMSE and PAE are significantly reduced after denoising for 7 and 6 kinematics components respectively ($p<0.05$, Wilcoxon signed-rank test). After denoising, the mean RMSE reduction averaged across different kinematics components reached 36% and the mean PAE reduction reached 56%. The largest reduction in the mean RMSE (relative to that of the original mouthguard kinematics measurements RMSE) reaches 56% (Z-axis linear acceleration) while the largest reduction in the mean PAE reaches 86% (X-axis angular velocity). On the 9-impact test subset, due to the small sample size, RMSE and PAE are reduced for 5 channels and 4 components with statistical significance ($p<0.05$), while the mean and median errors are reduced for all 8 components. To sum up, after denoising, on the kinematics level, the mouthguard kinematics measurement error is reduced when compared with the reference kinematics measurement from sensors implanted in the ATD.

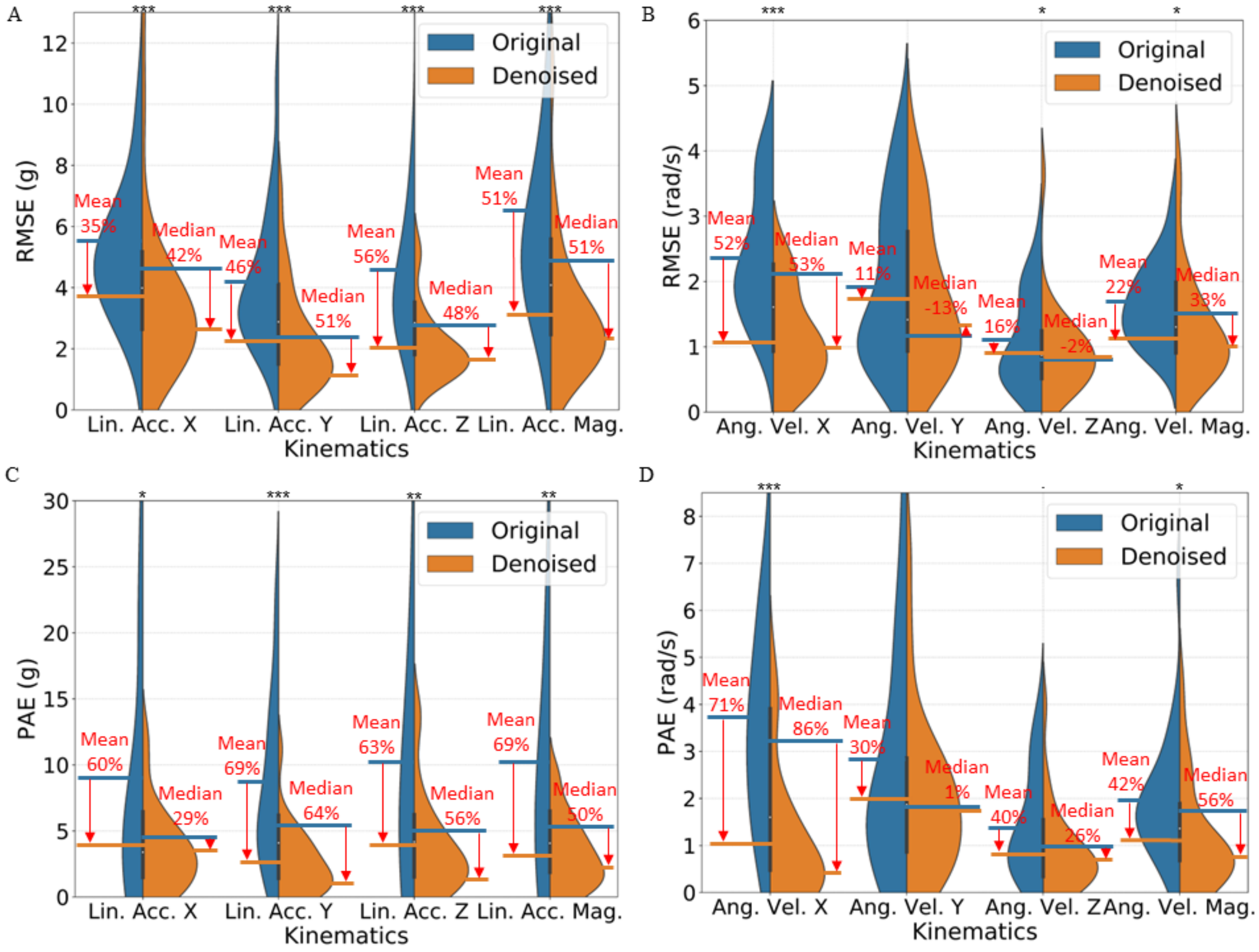

**Figure 3. The distribution of the pointwise root mean squared error (RMSE) and peak absolute error (PAE) between the reference kinematics and the mouthguard measurement with/without denoising on the 25 test impacts.** The reference kinematics (linear acceleration and angular velocity, each with tri-axial components and the magnitude) are from the ATD sensor measurement. The RMSE is calculated on each time point and the PAE is calculated at the peak value of the kinematics. The percentage reduction in the error median and the statistical significance between the errors from the original signals and the denoised signals are noted (.: $p<0.1$, *:$p<0.05$, **: $p<0.01$, ***: $p<0.001$, Wilcoxon signed-rank test).

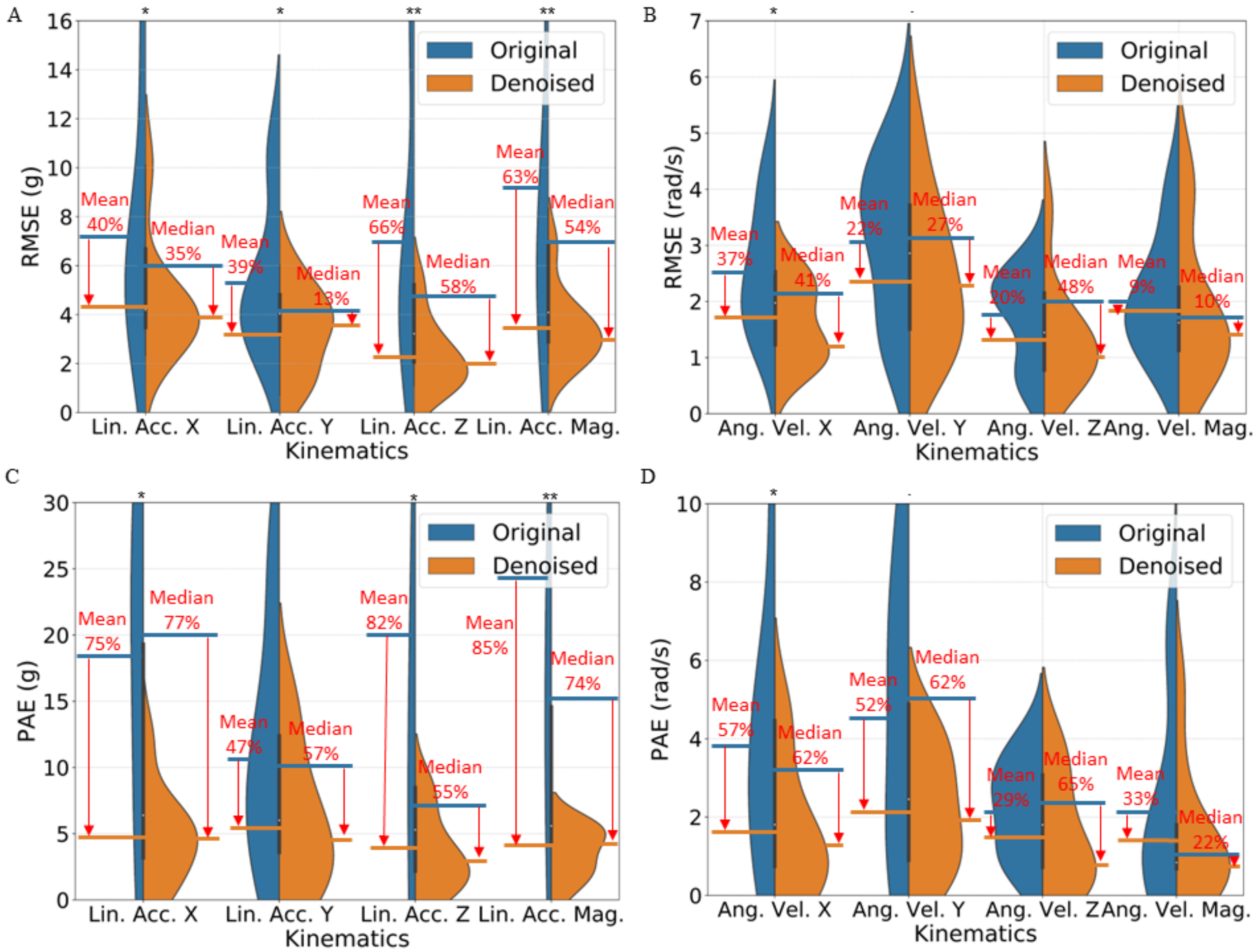

**Figure 4. The distribution of the pointwise root mean squared error (RMSE) and peak absolute error (PAE) between the reference kinematics and the mouthguard measurement with/without denoising on the 9-impact test subset.** The reference kinematics (linear acceleration and angular velocity, each with tri-axial components and the magnitude) are from the ATD sensor measurement. The RMSE is calculated on each time point and the PAE is calculated at the peak value of the kinematics. The percentage reduction in the error median and the statistical significance between the errors from the original signals and the denoised signals are noted (.: $p<0.1$, *:$p<0.05$, **: $p<0.01$, ***: $p<0.001$, Wilcoxon signed-rank test). The 9 test impacts from the no-repeat dataset are particularly selected to evaluate the model performance on completely unseen impact directions and impact velocities.

2. **Mouthguard denoising enables the brain injury criteria calculated based on the kinematics to be more accurate and lowers the risk of TBI risk misinterpretation.**

In addition to the comparison on the kinematics level, we further analyzed the effect of denoising on the brain injury criteria (BIC) level, again referencing the ATD as the gold-standard. BICs are the injury metrics developed by researchers via reduced-order modeling or statistical fitting to rapidly compute the risks of TBI [10,22]. In this study, we investigated six BICs: head injury criterion (HIC, [11]), head injury power (HIP, [27]), Generalized acceleration model for brain injury threshold (GAMBIT, [28]), severity index (SI, [29]), brain injury criterion (BrIC, [12]) and combined probability of concussion (CP, [30]). The equations to compute these metrics are shown in Method Section 4. We selected these BICs since they are well-established injury metrics which represent the linear-acceleration-based, angular-velocity-based and angular-acceleration-based BICs and they have been widely used in TBI risk evaluation in fields of automobile crashworthiness tests and helmet testing. It should be mentioned that more recently developed BICs have shown better correlation with brain strain [10,31] but we will proceed with the discussion on the tissue-level strain and strain rate level in the next section. Therefore, we will limit our study on the six BICs listed above. We calculated the absolute error between the BIC values computed based on the reference kinematics measurements and those computed based on the mouthguard kinematics measurements with/without denoising.

It is shown in Fig. 5 (25-impact test set) that on four out of the six BIC, denoising significantly reduced the error in the BIC calculation ($p<0.001$) while on BrIC and CP there was no statistical significance ($p>0.1$). The relative reduction in the mean error reduction was 65%. On the 9-impact test subset (Fig. S3), the mean error reduction on six BIC was 82%. To sum up, the denoising generally reduced the error in the calculation of selected BICs.

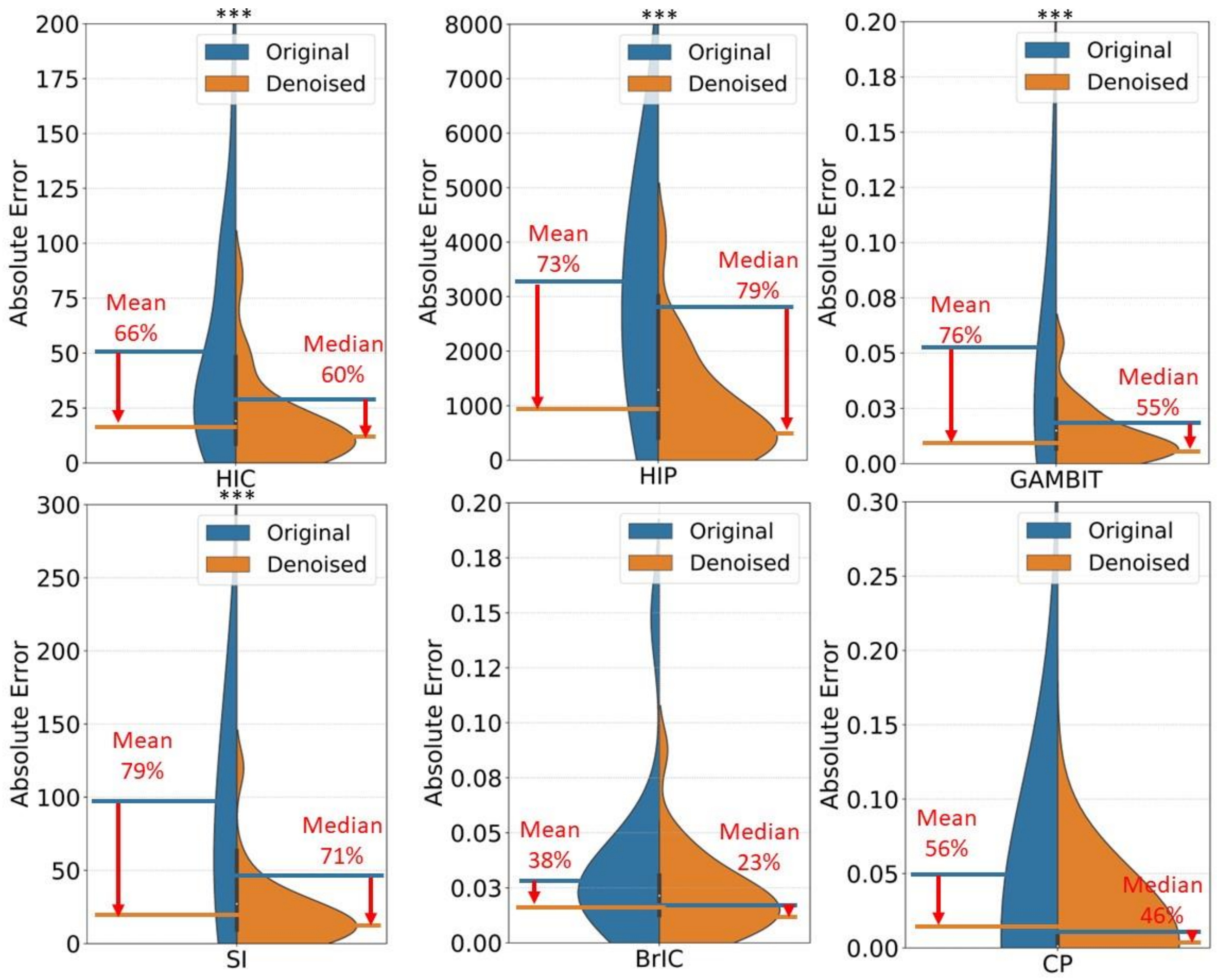

**Figure 5. The distribution of the absolute error between the BIC values calculated by the reference kinematics measurements and that calculated by the mouthguard kinematics measurements with/without denoising on the 25 test impacts.** The absolute error in six BIC values are shown. The reference BIC values are calculated by the ATD sensor kinematics measurements. The percentage reduction in the mean and median error and the statistical significance between the errors from the original signals and the denoised signals are noted in the plots (.: $p<0.1$, *:$p<0.05$, **: $p<0.01$, ***: $p<0.001$, Wilcoxon signed-rank test).

3. **Mouthguard denoising enables the tissue-level brain strain and strain rate calculated based on the kinematics to be more accurate for accurate whole-brain deformation estimation.**

The effects of denoising on the calculation of tissue-level brain strain and strain rate were further evaluated. We performed FEM to calculate brain tissue strain and strain rate based on the mouthguard kinematics measurement with/without denoising and compared the strain/strain rate results calculated from the ATD reference kinematics measurements. Since from the physiological aspect, the strain and strain rate are the key parameters for inertial brain injury, we deem that the denoising effect on the strain and strain rate level plays a crucial role in the injury risk estimation. The KTH FEM [32], which modeled 4,124 brain elements, was leveraged in this study to compute the whole-brain MPS and MPSR.

The element-wise absolute error in MPS and MPSR estimation is shown in Fig. 6. According to the results in Fig. 6 (A) and (B), the denoising led to significantly reduced error in MPS and MPSR estimations on both the entire test set and the 9-impact test subset ($p<0.001$). The statistics shown in Fig. 6 (C) indicate that the MPS and MPSR estimated by the denoised mouthguard kinematics measurements are more accurate than those calculated by the original mouthguard kinematics measurements: the mean, median and standard deviation in the MPS error and MPSR error were reduced on both MPS and MPSR on the two sets of test impacts. However, the extent of the denoising effect on the relative error reduction was not as evident as those on the levels of kinematics and brain injury criteria. To quantify the varying denoising effect across different brain regions, the reduction in MPS and MPSR estimation error across different brain regions is shown in Table 4 and Fig. S4. The results show that on the tissue-level strain and strain rate, the denoising effect is most evident in the brain regions such as the brainstem, cerebellum and midbrain, which indicates that the denoising effect varies with different brain regions.

To show how the denoising affects the concussion risk derived by strain and strain rate, we fitted two logistic regression models on the National Football League dataset [33] with 53 impacts (concussion: 22, non-concussion: 31) to predict the binary outcome of concussion based on the 95th percentile MPS and 95th percentile MPSR. Then, the strain and strain rate calculated based on the mouthguard kinematics were transformed into concussion risk (by taking the spatial 95th percentile and the logistic function) and compared with that calculated from the reference kinematics. The results on the entire test set and 9-impact test subset are shown in Fig. S5, which indicates that the concussion risk estimation was generally more accurate after denoising than without (error mean and standard deviation reduced). However, there was no statistical significance in the comparison of error distribution ($p>0.1$, Wilcoxon signed-rank test).

To summarize, on the level of tissue-level strain and strain rate, the denoising effectively reduces the error in the estimation of whole-brain element-wise MPS and MPSR. This implies that the denoising enables more accurate detection of high strain regions and high strain rate regions in the brain from head impacts measured with instrumented mouthguards.

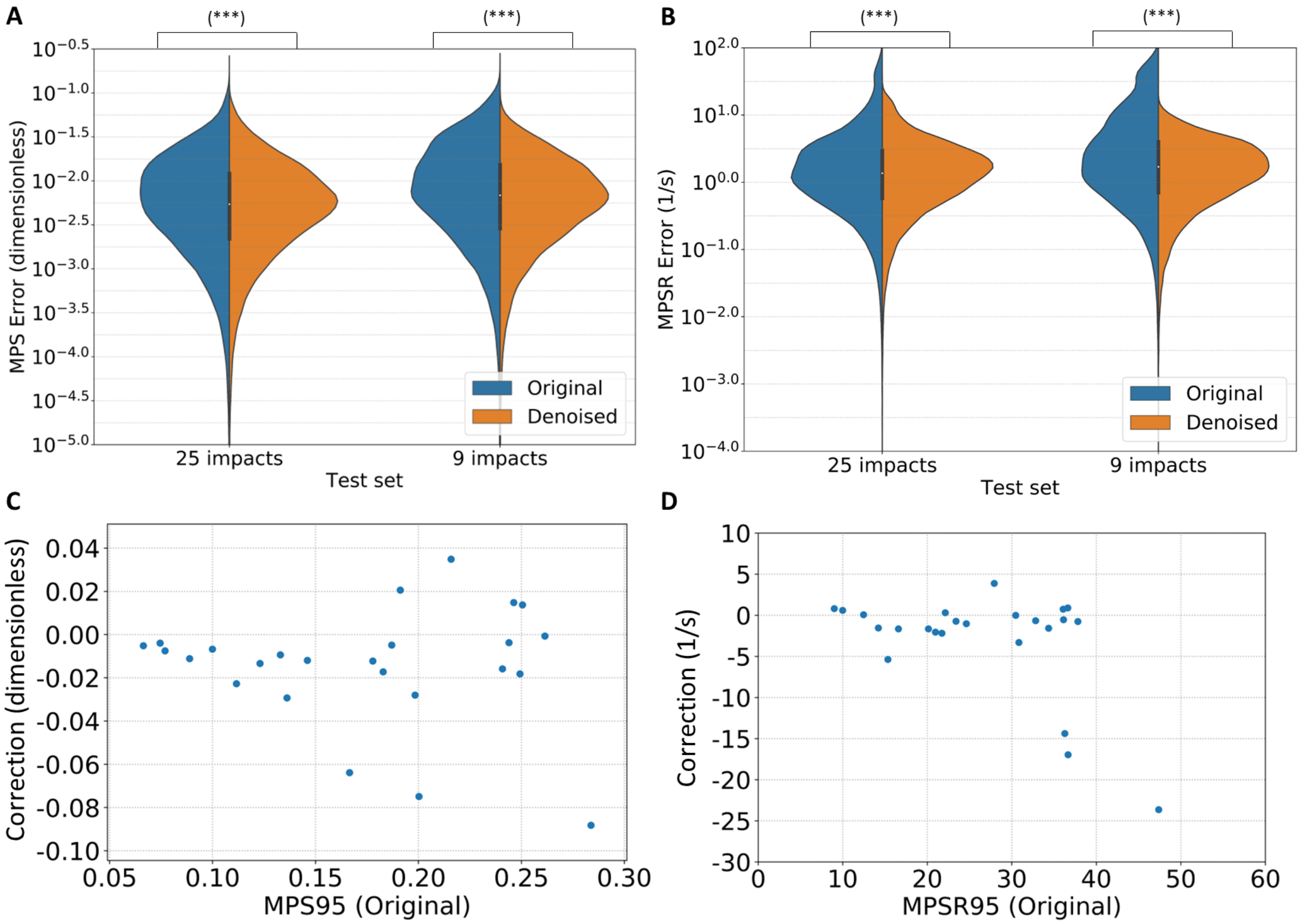

**E**

| Metric | Method | Test Set | Median | Mean | Standard Deviation |
|---|---|---|---|---|---|
| MPS Error | Original | 25 impacts | 0.006 | 0.010 | 0.013 |
| MPS Error | Denoised | 25 impacts | 0.005 | 0.009 | 0.010 |
| MPS Error | Original | 9 impacts | 0.008 | 0.014 | 0.017 |
| MPS Error | Denoised | 9 impacts | 0.006 | 0.009 | 0.010 |
| MPSR Error | Original | 25 impacts | 1.367 | 3.570 | 7.680 |
| MPSR Error | Denoised | 25 impacts | 1.361 | 2.213 | 2.755 |
| MPSR Error | Original | 9 impacts | 2.137 | 6.685 | 11.908 |
| MPSR Error | Denoised | 9 impacts | 1.453 | 2.081 | 2.007 |

**Figure 6. The denoising performance on the tissue-level strain and strain rate on the test impacts.** The log-scale violin plots of the element-wise absolute error in MPS estimation (A) and the MPSR estimation (B) on the 4,124 brain elements calculated with the KTH FE head model, the correction in 95th percentile MPS (C) and 95th percentile MPSR (D) led by denoising (denoised - original), and the statistics of the element-wise MPS/MPSR error (E). The reference MPS and MPSR values are calculated using the kinematics from the ATD sensor. The statistical significance between the errors from the original and denoised signals are noted in the plots (.: $p<0.1$, *:$p<0.05$, **: $p<0.01$, ***: $p<0.001$, Wilcoxon signed-rank test). The 9 test impacts from the no-repeat dataset are specifically selected to evaluate the model performance on completely unseen impact directions and impact velocities.

## 4. Denoising on the on-field football impacts leads to significantly different kinematics, BIC and brain deformation estimation.

We have shown that on the test impacts collected in the laboratory setting, the denoising of the mouthguard kinematics measurements based on the 1D-CNN models leads to more accurate kinematics, BIC estimations and whole-brain element-wise strain and strain rate estimations. To evaluate the effect of the denoising on real-world on-field impacts (the impacts measured by instrumented mouthguards worn by contact sport athletes participating in games), we applied the 1D-CNN denoising models on a blind test dataset of 118 impacts collected from collegiate football games [18,25]. The test on the 118 impacts was deemed a blind test because there were no reference kinematics (no implant kinematics sensors inside the players' heads). We calculated the difference in the magnitude of linear acceleration and angular velocity measurements, and in the whole-brain tissue-level MPS and MPSR (Fig. 7). The denoising with 1D-CNN models generally reduced peak kinematics, strain, and strain rate values. Furthermore, as can be seen from the distribution of the whole-brain element-wise MPS and MPSR in Fig. S6, denoising with 1D-CNN models significantly reduces the element-wise MPS and MPSR ($p<0.001$).

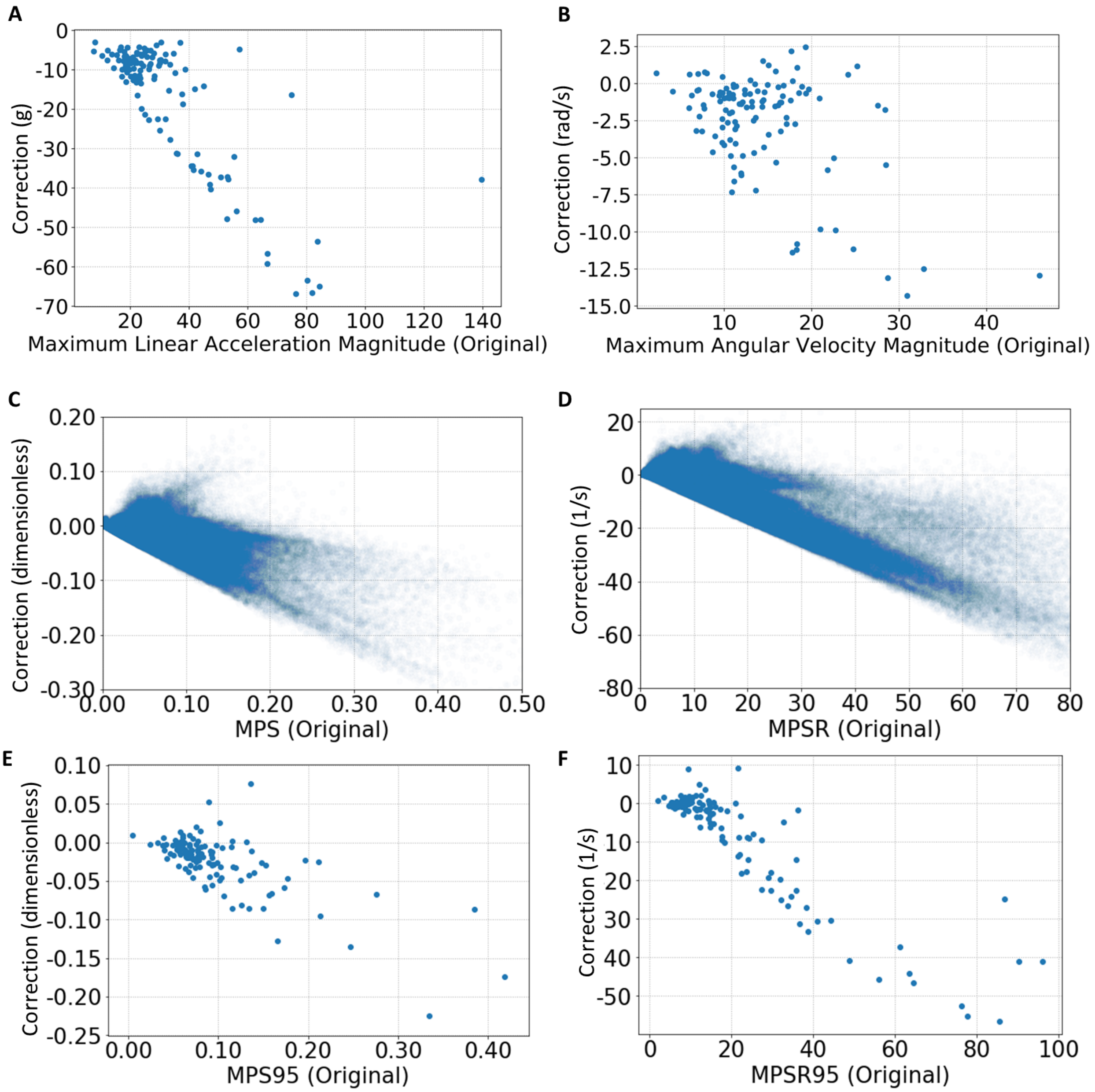

**Figure 7. The effect of denoising on the 118 on-field college football impacts for the blind test.** The x-axis denotes the values calculated based on the original signal of mouthguard kinematics measurement. The y-axis denotes the correction (values based on the denoised mouthguard kinematics measurements - values based on the original mouthguard kinematics measurements) in the maximum linear acceleration magnitude (A), the maximum angular velocity magnitude (B), whole-brain MPS (C), whole-brain MPSR (D), 95th percentile MPS (E) and 95th percentile MPSR (F) between the denoised kinematics measurement and the original signals of mouthguard measurements. Plots (C) and (D) show the results on 4,124 brain elements for each impact.

5. **1D-CNN models can attenuate noise caused by direct impacts to the mouthguard in post-mortem human subject impacts with unconstrained mandibles.**

As an exploratory experiment to test how the denoising models work in a worst-case scenario, we applied the 1D-CNN models to data collected from 413 impacts to post-mortem human subjects (PMHS). In these impact tests, a completely unconstrained mandible could move freely and strike the mouthguard during the impact. Part of these data were previously used in a study to quantify the error caused by the direct impact on mouthguard by Kuo et al. [53], which found that the normalized RMSE can reach 40-80%.

The denoising effect and the errors of kinematics measurement are shown in Fig. 8. It can be seen from Fig. 8 (A) that the denoising effect in the PMHS dataset was generally similar to the effect on the on-field head impact data for the college football players shown in Fig. 7 (A): the peak kinematics were generally reduced after the denoising and the reduction is more evident as the absolute value of the maximum kinematics peak increases. According to the statistics shown in Fig. 8 (B), when compared with the reference kinematics (measured by sensors screwed into the posterior part of the skull of the PMHS), the denoised kinematics were more accurate when compared with the original mouthguard kinematics, with increasing Pearson correlation coefficient and decreasing RMSE. When the top 10% and top 5% noisiest impacts were investigated, according to the results shown in Fig. 8 (C) and (D), the denoising effect was evident: the relative reduction of the absolute error in maximum angular velocity magnitude reached 75.6% and 82.3% respectively. Averaged over the entire test set, the improvement in the kinematics was not as evident as in the standard protocol and no-repeat datasets where the mandible was fixed open. Our results demonstrate that for this worst-case scenario (a completely unconstrained mandible and a direct impact to the mouthguard), the denoising models can slightly attenuate the noise despite the denoising models not being designed for this scenario and this type of data not being included in the training dataset.

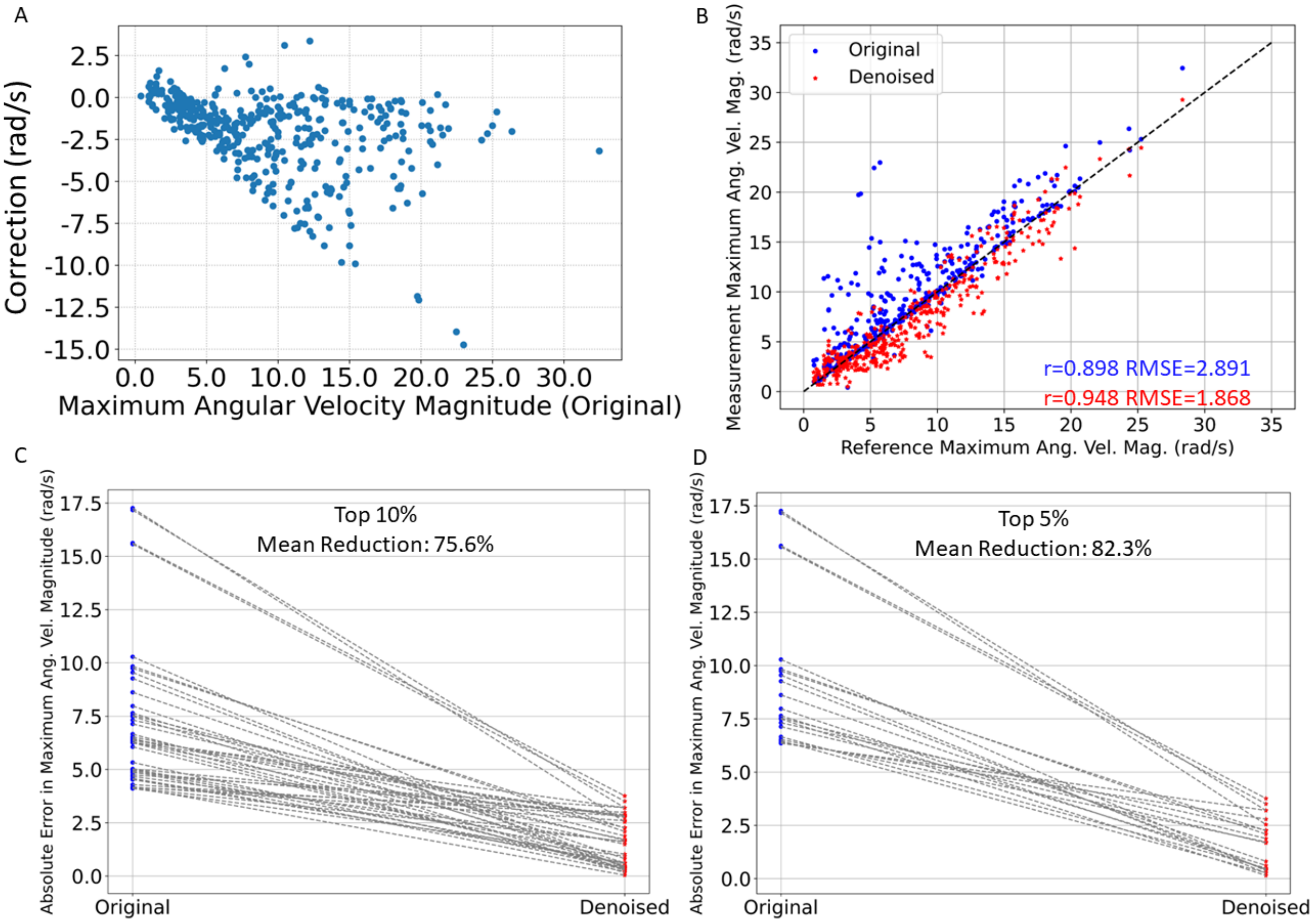

E

| Channel | RMSE | Peak_RMSE | Peak $R^2$ | Peak_r |
|---|---|---|---|---|
| Ang. Vel. X (original) | 6.646 | 3.845 | 0.151 | 0.661 |
| Ang. Vel. X (denoised) | 5.823 | 3.626 | 0.245 | 0.689 |
| Ang. Vel. Y (original) | 4.096 | 4.585 | -0.198 | 0.580 |
| Ang. Vel. Y (denoised) | 3.963 | 3.392 | 0.344 | 0.646 |
| Ang. Vel. Z (original) | 4.978 | 1.776 | 0.674 | 0.826 |
| Ang. Vel. Z (denoised) | 5.297 | 1.792 | 0.668 | 0.846 |
| Ang. Vel. Mag. (original) | 0.905 | 2.891 | 0.702 | 0.898 |
| Ang. Vel. Mag. (denoised) | 1.499 | 1.868 | 0.876 | 0.948 |

**Figure 8. The denoising effect and the error in kinematics measurements on the 413 post-mortem human subject head impacts.** The denoising effect is shown in (A) by the correction (values based on the denoised mouthguard kinematics measurements - values based on the original mouthguard kinematics measurements) in the peak magnitude of angular velocity. (B) shows the peak angular velocity magnitude of the reference, original mouthguard measurements, denoised mouthguard measurements. (C) and (D) shows the reduction of absolute error in maximum angular velocity magnitude among the top 10% and top 5% noisiest impacts. (E) shows the error in the kinematics measurements with/without denoising.

## 6. 1D-CNN models are more effective in denoising mouthguard kinematics than a simple signal filtering approach.

It has been shown from our blind-test results that 1D-CNN denoising led the peak kinematics and tissue-level strain/strain rate to be reduced. To ascertain whether additional filtering works better than the deep learning approach, we compared the denoising with signal filtering based on a low-pass Butterworth filter (orders: 5-9, cutoff frequencies: 20-160Hz) and the results are shown in Fig. S7, S8 (metrics shown include: MAE, RMSE, Peak-MAE, Peak-RMSE, Peak-$R^2$). These results reveal that, generally, with a large advantage, the 1D-CNN models outperformed the simple signal filtering approach in both pointwise and peak-wise comparisons with the reference kinematics measured by the ATD sensors (exceptions on three of the six metrics on linear acceleration along Z-axis). These results indicate that the data-driven denoising approach with 1D-CNN models is more effective than simply filtering high-frequency components in generating kinematics signals better correlated with the reference ATD sensor kinematics.

**Discussion**

Based on laboratory head impacts collected from an ATD headform impacted by a linear pneumatic impactor, we developed a set of 1D-CNN models to denoise the kinematics measurements recorded by instrumented mouthguards. This led the denoised kinematics measurements to be closer to the reference kinematics measurements recorded by the sensors implanted inside the ATD head. The effectiveness of denoising was evaluated on three levels: the kinematics, the brain injury criteria, and the tissue-level strain and strain rate calculated based on the kinematics. The effectiveness of denoising was also validated on a test subset with impact conditions completely unseen in the training dataset. The 1D-CNN models were also tested on 118 on-field college football impacts and the results showed the models contributed to statistically significantly lower tissue-level strain and strain rate. As for the implications, these models can potentially be used on in-vivo data and improve the quality of head impact data.

As the brain is damaged due to the inertia of the brain during the rapid movement of the head, accurate measurements of head kinematics may enable researchers and clinicians to better detect dangerous head impacts that can lead to concussion and multiple TBI pathologies including traumatic axonal injury and blood-brain barrier disruption. Over the past decades, multiple kinematics measurement technologies for head impacts have been developed. Instrumented mouthguards have shown the ability to measure the head kinematics that better correlate with ATD sensor measurements when compared with other technologies [34]. However, during real-world head impacts, the loosening of the instrumented mouthguard and the electronic noise can interfere with the accurate measurement of real head kinematics. For example, Liu et al. [18] has shown that on different types of instrumented mouthguards, there can be an average of 2.5 to 32.4% relative error in peak linear acceleration and angular velocity, respectively. To further improve the accuracy of the kinematics measurements, the 1D-CNN models aim at predicting the reference kinematics measurements recorded by the ATD sensors with the original mouthguard kinematics measurements as the input, and this is the first work on denoising instrumented mouthguard's kinematics measurements to the best of our knowledge. Efforts have been put into improving the hardware of measurement technologies, and this study presents an AI-based model to improve the measurement accuracy for the first time. In signal processing, low-pass filters were used to improve the accuracy of the measurement. It should be noted that the filters only remove the high-frequency component in the data, while the 1D-CNN models also address the deviation in low-frequency components and can sometimes supplement components. In this paper, both the original mouthguard data and the reference data were both filtered by optimized filters according to our previous study [18].

As for the efficacy of the AI-based denoising, we have shown that after denoising, the kinematics measurements originated from instrumented mouthguards better match the reference kinematics originated from the ATD sensors. The errors in peak kinematics and the pointwise root mean squared error can be reduced by approximately 80%. As for the implication of the AI-based denoising, with further model validation across different impact datasets and different types of mouthguards, the 1D-CNN denoising models can enable researchers to clean the sensor measurements and measure the head kinematics more accurately.

More importantly, it has been shown that after denoising, the errors in the estimation of brain injury criteria and the associated estimated concussion risks based on the logistic regression injury risk functions can also be significantly reduced. Because the BICs are directly computed based on the kinematics via mathematical transforms, the denoising on kinematics has a direct effect on the BIC estimation. Furthermore, with access to FEM, the denoised mouthguard kinematics measurements enable more accurate estimation of tissue-level strain and strain rate across the entire brain. Different from the BICs which can only provide one value for the entire brain, the distribution of the whole-brain strain and strain rate can be computed. Therefore, different from the injury risk estimation based on the BICs, the whole-brain strain and strain rate provides the most direct evidence of regional and tissue-level injury risks.

In the past, the region-specific strain and strain rate information has been shown to be predictors of the blood-brain barrier disruption [20] and traumatic axonal injury linked to specific brain regions [35]. For example, O'Keefe et al. [20] has shown that the high strain regions manifest high Pearson correlation with the brain regions with blood-brain barrier disruption evidenced by dynamic contrast enhanced magnetic resonance imaging. Hajiaghamemar et al. [35] have shown that on a large animal model, the region-specific strain and strain rate are good predictors of the location of the traumatic axonal injury. As a result, the improved accuracy in the tissue-level strain and strain rate estimation resulting from the denoised mouthguard kinematics can provide the users of the instrumented mouthguards with the high-dimensional region-specific biomechanics loadings for each impact detected. Once a dangerous impact occurs, for example, when high strain and strain rate at vulnerable brain regions such as the corpus callosum are detected, real-time interventional approaches and further medical diagnostics suggestions can be provided to the sufferers of the head impacts.

The potential noise that has been removed by the 1D-CNN models is also worth discussion. Mouthguards are fitted to the upper dentition by the frictional force between the teeth and the mouthguard surface. To increase this frictional force, the mouthguard was customized to enhance the compression and contact area between the mouthguard and the teeth. Considering the comfort of the mouthguard, soft and elastic materials are preferred to make the mouthguard, so it is possible that the fit between the mouthguard and the dentition gets loose, potentially leading to relative movement, which in turn leads to noise and errors in the measurements. These suggested that the measurement error depends on the characteristics of the impact, and the error can be estimated by developing a model based on the kinematics. In the history of instrumented mouthguard development, the measurement accuracy was tested in different methods: the mouthguard was clenched in [53], which prohibited any relative movement and yielded a high accuracy. Then a spring-articulated mandible was included in headform to test the instrument mouthguard, and the accuracy was extremely low because of the impacts from the spring-driven mandible [49]. The different methods to fix mouthguard was investigated in [40], and the impacts from mandible were avoided in the further studies [48,18]. The testing method in [18] was used in this study, so noises caused by the mandible impacts were not suppressed by this model. Additionally, the intrinsic electronic noise of the accelerometers and gyroscopes themselves can also contribute to the inaccuracy of mouthguard kinematics. In previous studies [18], errors of measurement were found to vary among the impact locations and a similar tendency can be observed: the mouthguard measurement in the front and back impact locations are always higher than the reference kinematics, and the measurement in the side impact has the highest accuracy [18]. The data in [18] was used in this study as the standard protocol dataset, and we noticed that the noise in the no-repeat dataset was higher than the standard protocol dataset. Although we used the same instrumented mouthguard to collect the data, it is possible the aging of mouthguard material loosened that fit between the mouthguard and the dentition, which produced more relative movement. Introducing the no-repeat dataset diversified the noise level in the training dataset, which improved the model's ability to identify and attenuate noise. Further, it should be noticed that determination coefficients between the mouthguard readings and the reference reported in [18] was different from pearson correlation presented in Fig.2. Therefore the value should not be directly compared.

The denoising effect on the kinematics from post-mortem human subject impacts with unconstrained mandibles and direct impacts to the mouthguard is also worth discussion. Based on the results of the PMHS data [53], it can be shown that the denoising can lead the kinematics to be more accurate with regard to the reference sensor measurements. Particularly, for the top 10% noisiest data and top 5% noisiest data with the highest kinematics error, the 1D-CNN models were very effective as is shown by the over 75% relative reduction in the absolute error in maximum angular velocity magnitude in Fig. 8 (C) and (D), which indicates that the denoising models can be effective in compensating the extremely high noisy in the kinematics. According to the previous publication [53], the PMHS data represents the worst-case scenario where the mouthguard is impacted directly and moves independently from the upper skull. Significant noise (normalized RMSE of 40-80%) was produced due to the mouthguard being dislodged from the upper dentition during impact. Although the 1D-CNN denoising models were not developed for this extreme case (because the training data of the 1D-CNN consisted of impacts collected from a headform with a spring-articulated mandible fixed open) and the head kinematics have different characteristics (training data was collected with a Hybrid III neck, while the PMHS data was collected without a neck), the performance of these denoising models show that it can still attenuate the noise under this condition. It can be observed from Fig. 8 (A) and (B) that on the PMHS dataset, the denoising is effective in reducing the high peak angular velocity magnitude caused by the head impacts, which can be helpful in addressing the problems of false positives in the detection of severe/concussive head impacts to improve the accuracy and precision of dangerous head impact detection.

Additionally, it should be noted that the denoising effect on the PMHS data showed similar patterns observed in the on-field blind test data: the kinematics peaks are reduced and the changes increase with the absolute value of the kinematics peaks. Therefore, it is possible that the 1D-CNN models can improve head impact data collected in settings where direct mouthguard impacts frequently occur (e.g., combat sports). However, more comprehensive studies are needed to determine this.

The usage of 1D-CNN over alternative approaches also merits further discussion. The 1D-CNN leverages the temporal correlation and performs the one-dimensional convolution along the axes of time for each channel of the kinematics signals. Therefore, the shape of the kernels and the receptive fields are rectangular with width 1. As a result, the convolution is able to detect local patterns in a sliding time window (the convolution kernel) and leverages the local information and the patterns detected to denoise the mouthguard kinematics measurements. We chose the 1D-CNN structures because they are efficient and better at mitigating overfitting to the training data and shortening the computational time when compared with the recurrent neural network. The complicated structure of a recurrent neural network cell entails more computations that may lead to overfitting and lengthy computation. Additionally, the 1D-CNN's local convolution strategy is better at flexibly detecting temporal patterns than the fully connected neural network.

In a fully connected neural network, the model matches certain time points with specific impact-related information, which is less reliable because the actual time when the impact kinematics traces start, the peak and decay can vary widely among the impacts. This also corresponds well with the real-world scenarios: as the types of head impacts vary significantly (e.g., American football, MMA, car crashes) and the duration and frequency components of different types of head impact kinematics vary largely [36], the time traces of the impact kinematics can be significantly different. Even for the impacts from the same type, say American football, the impacts can have single peak or multiple peaks, the trace can also bend upward or downward along each axis due to different impact locations, and therefore, the time traces of kinematics can appear significantly different. As a result, the fully connected neural networks' strategy of matching each time point's signal value to an input node may not be very effective in the actual denoising. This may explain why the 1D-CNN's local convolution works well under the highly variable time traces of the kinematics measurements.

Although this study provides effective denoising results with 1D-CNN models, there are several limitations that are worth discussing. First, the dataset may contain bias because the linear pneumatic impactor in the laboratory is designed to simulate impacts sustained in American football. The impacts generated by the system may not be representative enough of the real-world impacts seen in other sports [37]. In the future, multiple datasets of different types of head impacts can be developed by biomechanics researchers in the respective field to further validate the 1D-CNN models' performance across different types of impacts or develop different sets of denoising models for different types of head impacts.

Second, we leveraged the KTH finite element model in the calculation of the strain and strain rate, which is a validated finite element model for brain biomechanics modeling after head impacts. However, when compared with the recently developed finite element head models [38,39], the KTH model is limited with the absence of the gyri and sulci as well as the cerebral vasculature. The recently developed models are capable of calculating axonal strain and axonal strain rate, which have shown better correlation with traumatic axonal injury in large animal models [6]. Therefore, the denoising effect on the strain and strain rate shown in this study is limited to MPS and MPSR. In the future, more complicated finite element models can be used to further validate the effect of mouthguard denoising on the tissue-level axonal strain and strain rate.

Third, there is an absence of real-world reference kinematics measurements to evaluate the denoising effect in on-field impacts. This limitation can be a barrier to the validation of the potential applications because implanting kinematics sensors in human subjects is not only invasive but also requires the craniectomy, which imposes health threats to the human subjects and is associated with ethical concerns. Therefore, the lack of real-world reference kinematics measurements can be a limitation that is hard to address for this type of research.

Fourth, the noise caused by direct impacts to the mouthguard is not considered in this study. When the athlete's mouth is impacted directly, the impact force will propagate to the mouthguard sensors, potentially causing a large amount of error. We did not include this type of error because of the difficulties in collecting a labeled training dataset. It would be necessary to model the lip mechanical properties to include the noise caused by direct impacts to the mouthguard sensor. We applied the denoising model on PMHS data where the mandible was not fixed, resulting in a direct impact to the mouthguard from the lower dentition. However, the PMHS data was collected without a neck, so the head kinematics are not a suitable representation of real head impacts. To address this issue, realistic PMHS impacts that include a neck should be performed. The resulting data can then be merged into the training dataset.

Furthermore, in this study, we developed the models to denoise the Stanford instrumented mouthguards but we did not test the models on other types of instrumented mouthguards [18]. In the future, the model performance can be further validated on other types of mouthguards. If the model performance declines, additional transfer learning and domain adaptation can be added to the current 1D-CNN models to better fit different types of mouthguards.

Additionally, other deep learning approaches such as transformers have shown high effectiveness in temporal data mining and time-series applications [41,42]. First developed in the field of natural language processing, with various attention mechanisms, transformers are capable of detecting varying local and global patterns in the temporal signals. Their potential may be tested in the denoising of mouthguard kinematics measurements. In the future, denoising with transformers also warrants further investigation.

## Conclusion

In this study, we developed a set of one-dimensional convolutional neural networks to denoise the kinematics measurements of instrumented mouthguards. Trained on a dataset augmented from 113 head impacts collected with an anthropomorphic test headform and linear pneumatic impactor, the models showed high effectiveness in denoising the mouthguard kinematics measurements in the test set unseen in the training process. When compared with reference kinematics measurements, denoised mouthguard measurements showed significantly reduced errors in peak kinematics, brain injury criteria and tissue-level strain and strain rate. The denoising also reduced the strain and strain rate for a blind-test dataset of on-field college football impacts. This study provides users in the fields of impact biomechanics and traumatic brain injury with a tool to improve the accuracy of the measured head kinematics for TBI risk estimation and prevention of brain damage accumulation after dangerous head impacts are detected.

## Method

### 1. Experiment setup

The datasets used in this study were collected from a laboratory head impact setup: a Mandible Load Sensing Headform (MLSH) ATD with a Hybrid III neck was mounted to a plate sliding along the impact direction. The impact location can be adjusted by rotating the ATD in sagittal and horizontal planes and moving the plate. The impact loading was delivered by a ram which is accelerated by air pressure. The ATD wore a football helmet properly. A set of high-accuracy accelerometers and gyroscopes were mounted at the center of gravity (CoG) of the headform, and the reading was used as the reference head kinematics. The MLSH has a high-biofidelity upper detention, where we fitted the Stanford instrumented mouthguard on (Boil-and-bite and customized in [18]). In each impact, the mouthguard readings were filtered with a cut-off frequency of 160 Hz, and then transformed to the CoG. Since the mouthguard and the ATD sensors were not triggered together, the mouthguard readings were shifted to synchronize with the ATD readings. To avoid the direct impact between the lower dentition and the mouthguard, the mandible was fixed open. Details of the impact testing setup and data processing are available in [18].

In the standard protocol dataset, four impact speeds (3.6, 5.5, 7.4, 9.3 m/s) and five impact locations (facemask, front, oblique, side and back) were included and three repetitions were performed for every combination of impact velocity and location in 2019. Then, to further test the denoising model with a dataset without repetition, we collected the no-repeat dataset in 2021 by impacting the head at front and side by rotating the head horizontally every 22.5° from 0° to 180°. Three impact speeds (4.2, 6.1, 7.5 m/s) were adopted. In the no-repeat dataset, we only used the Boil-and-bite mouthguard [18]. An example impact was shown in Video S1.

## 2. One-dimensional convolutional neural network denoising models

Convolutional neural network (CNN), developed based on the structure of human visual perception system and widely used in the field of computer vision, has been shown effective in signal denoising tasks in previous studies [23,24]. CNN is powerful in finding patterns in images and signals via the convolutional operation, which automatically extracts the features from the data in the model training process. Different from the conventional two-dimensional CNN, to apply the CNN for signal denoising, we developed the one-dimensional CNN (1D-CNN) models to extract the features from the temporal traces of the kinematics signals measured by sensors. As a result, we developed six 1D-CNN models for the six different components of kinematics (linear acceleration along the x-, y- and z-axes, angular velocity along the x-, y- and z-axes). For each model, the 1D-CNN model input was the original signal of the kinematics measured by instrumented mouthguards (after signal filtering) while the 1D-CNN model output was the reference signal of kinematics measured by the sensors in the ATD. It should be mentioned that all the original and reference signals are filtered.

We designed the model to be fully convolutional with six layers (visualized in Fig. 8): the first three layers are convolutional layers with increasing numbers of channels for sake of feature extraction, while the last three layers are transposed convolutional layers with a symmetric structure with the convolutional layers for sake of signal upsampling. The number of channels in the convolutional layers was deemed as a hyperparameter to be tuned on the validation data. The convolutional kernel size was also tuned as a hyperparameter. Rectified linear units (ReLU) was used as the activation function except after the last transposed convolutional layer (the output layer). The rationale of this designed structure was to firstly extract information from the one-dimensional time signals with an increasing number of filters and channels and then upsample the condensed information to output the one-dimensional denoised signals. Adaptive moment estimation (ADAM) optimizer was used to optimize the weights in the convolutional kernels via back propagation and the initial learning rate was tuned as a hyperparameter [43]. Additionally, based on the decreasing tendency of the validation loss and training loss, we further decayed the learning rate by a factor of 0.9 after each 50 training epochs to enable a smoother convergence of the loss. Early stopping was also performed based on the trend of training loss and validation loss to avoid model overfitting. Mean squared error was used as the loss function in this study. To further avoid overfitting, L2 regularization on the weights of the convolutional kernels were added to the loss function. It should be noted that we have also experimented with the recurrent neural networks and fully connected neural networks but the model performance was not as good as the 1D-CNN models' performance. Therefore, we reported the 1D-CNN models for mouthguard kinematics denoising in this study.

To train the model, we partitioned the entire dataset (163 impacts collected in the laboratory) into a training set (113 impacts, 70%), validation set (25 impacts, 15%) and test set (25 impacts, 15%). To enable more accurate mouthguard kinematics denoising models, we performed training data augmentation by using a sliding window (width: 100 ms, stride: 5 ms) to augment the quantity of training data. After the data augmentation, the 113 200-ms training impacts generated 2,260 100-ms training impacts. In the model evaluation stage, we truncated validation impacts and test impacts to be 100-ms impacts to match the length of the augmented training impacts. The reason to choose 100 ms as the window width was because of the observation that the duration of most real-world head impacts (i.e., MMA, college football) is shorter than 100 ms and for any impact with duration longer than 100 ms, they can be predicted by frames of 100 ms.

The hyperparameters tuned based on the pointwise RMSE on the validation impacts in this study include the number of channels of the convolutional layers, the initial learning rate, the number of training epochs, the strength of the L2 penalty and the kernel size. In addition to directly modeling the reference kinematics measurements as the output of the 1D-CNN models, we also designed a noise-subtraction strategy in training: we modeled the noise between the reference kinematics measurements and the original mouthguard kinematics measurements as the 1D-CNN model output while the original mouthguard kinematics measurement was the 1D-CNN model input. Then, we subtracted the predicted noise from the original mouthguard kinematics measurements in a pointwise manner. The assumption was that the noise is pointwise additive in the kinematics and this noise-subtraction strategy may be able to let the 1D-CNN models to focus on the modeling of the noise rather than the entire traces of the reference kinematics signals. We tested both the direct modeling strategy and the noise-subtraction strategy and evaluated their performance on the validation set. Then, for each kinematics component, the best strategy with the best hyperparameters was selected as the model and the performance was finally evaluated on the hold-out test impacts. The results of the best set of hyperparameters are shown in Table S1.

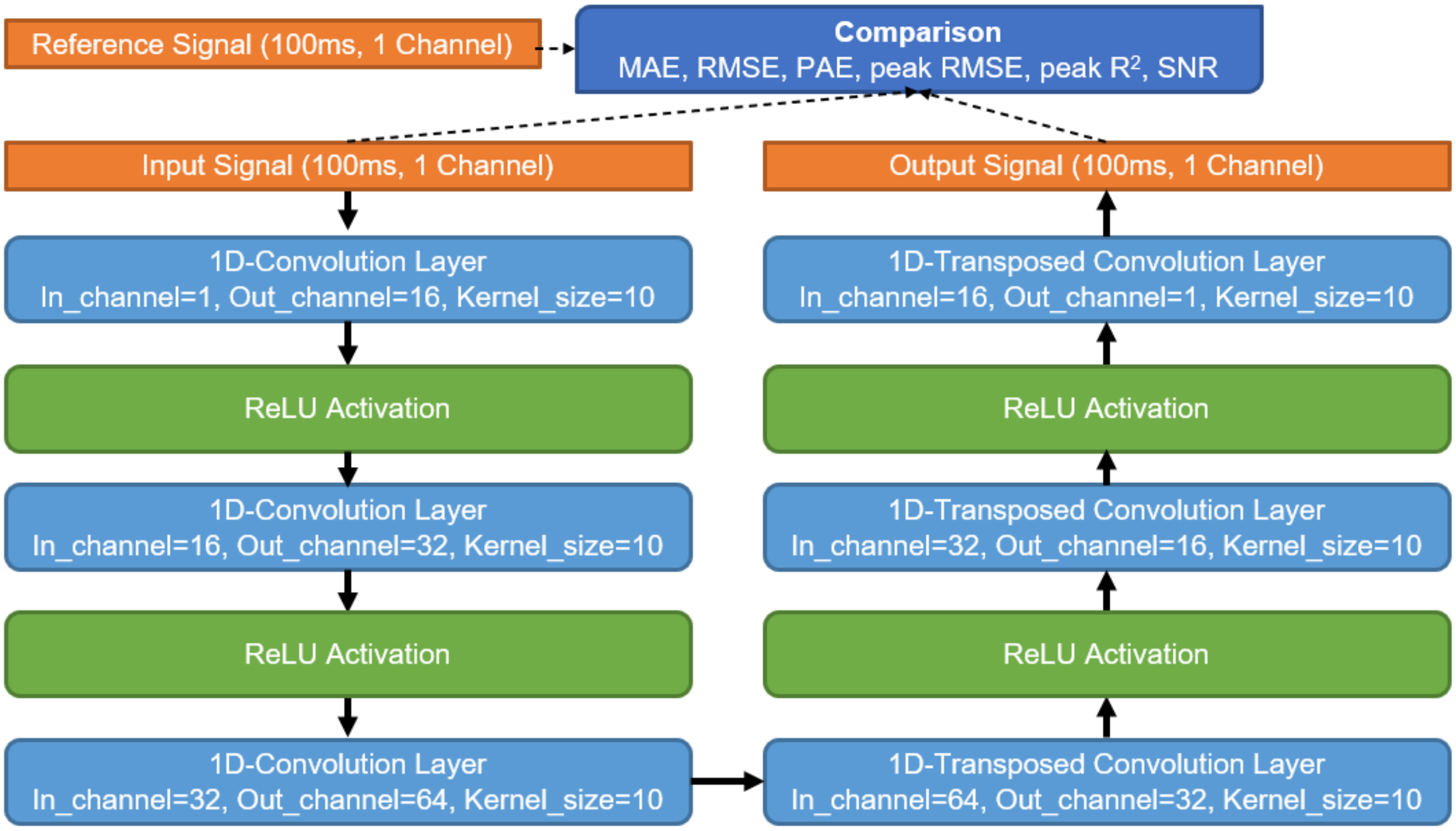

**Figure 9. The structure of the fully convolutional neural network kinematics denoising model.** Example is based on the best hyperparameters for denoising the X-axis of linear acceleration.

## 3. Brain injury criteria calculation

BIC are kinematics-based variables that were developed by researchers to rapidly quantify the risks of TBI. Different from the strain and strain rate, which rely on the time-consuming FEM based on whole-brain biomechanics and material properties, the BICs are directly calculated based on the kinematics with definite mathematical formulas and therefore can be computed in a flash, which can be used to quickly quantify the TBI risks. To evaluate the effect of mouthguard denoising beyond the kinematics level, we firstly evaluated the BIC calculated based on the mouthguard kinematics measurements with/without denoising. In this study, six different BIC were calculated: head injury criterion (HIC, [11]), head injury power (HIP, [27]), generalized acceleration model for brain injury threshold (GAMBIT, [28]), severity index (SI, [29]), brain injury criterion (BrIC, [12]) and combined probability of concussion (CP, [30]). These BIC were selected because they have shown their effectiveness in assessing TBI risk across different fields including vehicle crashworthiness tests and contact sports. These BIC are briefly introduced below:

The HIC, which is the most widely used BIC, was developed by Versace in the field of motor vehicle regulation [11]. The HIC is based only on linear acceleration and is calculated with the following equation (constraint: $t_2 - t_1 \leq 15ms$):

$$\text{HIC} = \max_{t_1,t_2}\left\{\left(\int_{t_1}^{t_2} |a(t)|dt\right)^{2.5} (t_2 - t_1)\right\}$$

The HIP was developed by Newman et al. based on the assumption that the head impact power correlates with the severity of the head impact [27]. The HIP is based on both linear acceleration and angular acceleration and is calculated with the equation below (Iii denotes the principal moments of head inertia while i denotes the kinematics component (along x-, y-. z-axes), m denotes the mass):

$$\text{HIP} = \max_{t}\left\{m \sum a_i(t) \int a_i(t)dt + \sum I_{ii}\alpha_i(t) \int \alpha_i(t)dt\right\}$$

The GAMBIT was developed by Newman et al. [28] to combine the translational and rotational components of head kinematics. The calculation of GAMBIT is based on the following equation (model constants: n=m=s=2, $a_c$=250g, $\alpha_c$ = 25000 rad/s$^2$):

$$\text{GAMBIT} = \left\{\left[(\frac{|a(t)|}{a_c})^n + (\frac{|\alpha(t)|}{\alpha_c})^m\right]^{1/s}\right\}$$

The SI was developed by Gadd [29] and is also known as the Gadd Severity Index. With the skull fracture data in head form simulation, the SI was based on the curve fitting of the Wayne State Tolerance Curve. The SI is based on linear acceleration. The calculation of SI is based on the following integral, where the integral starts when the signal firstly exceeds 4g and ends when the signal returns to 4g after the highest peak:

$$\text{SI} = \int |a(t)|^{2.5}dt$$

The BrIC was developed by Takhounts et al. [12]. It is based only on angular velocity, with the assumption that in pendulum and occupant crash tests the FEM-derived strains can be predicted by angular velocity alone. It is calculated based on the following equation (model constants: $[\omega_{xcr}, \omega_{ycr}, \omega_{zcr}]$, [66.2, 59.1, 44.2]):

$$\text{BrIC} = \sqrt{(\frac{\omega_x}{\omega_{xcr}})^2 + (\frac{\omega_y}{\omega_{ycr}})^2 + (\frac{\omega_z}{\omega_{zcr}})^2}$$

The combined probability of concussion (CP) was developed by Rowson and Duma [30], which has been widely applied in the rating of the football helmets. The calculation of CP is based on the concussion risk in football head impacts with a logistic injury risk function fitted to an injury dataset. It can be calculated according to the following equation (model constants: $\beta_0$ = -10.2, $\beta_1$ = 0.0433, $\beta_2$ = 0.000873, $\beta_3$ = -0.00000092):

$$\text{CP} = \beta_0 + \beta_1 \max_t|a(t)| + \beta_2 \max_t|\alpha(t)| + \beta_3\max_t|a(t)|\max_t|\alpha(t)|$$

**4. Finite element modeling for brain strain and strain rate calculation**

Physiologically, the brain is damaged because of the tissue-level deformation caused by the inertial movement of the brain after the rapid head rotation and acceleration. Strain and strain rate are metrics used to describe the severity of the brain deformation and have been shown to be good predictors of concussion and multiple TBI pathologies including blood-brain barrier disruption and traumatic axonal injury [21,31,35]. The state-of-the-art approach to compute the tissue-level strain and strain rate is through brain-physics-based FEM. In this study, we applied the KTH FE head model (Stockholm, Sweden) to calculate the tissue-level brain strain and strain rate. The KTH model is a validated FE model for mTBI-level head impacts. It models the entire brain with 4,124 brain elements which involve the brain, skull, scalp, meninges, falx, tentorium, subarachnoid cerebrospinal fluid, ventricles and 11 pairs of the largest bridging veins. The model has been validated by the fact that the responses of the head model correlated well with the brain-skull relative moderation, intracranial pressure and brain strain data acquired by experiments [44,45]. The modeling process was done in LS-DYNA (Livermore, CA, USA). At each time point, the kinematics were applied to the skull except that at the initial time point, the measured kinematics were applied to all elements of the brain.

**5. Concussion risk estimation based on strain and strain rate**

To investigate how the denoising on the kinematics affects the concussion risk calculated by the strain and strain rate, we fitted the logistic regression models on the National Football League impact dataset (53 impacts, concussion: 22, controls: 31) to predict the concussion based on the 95th percentile MPS and 95th percentile MPSR [33]. It should be mentioned that the 95th percentiles are typically used in the FEM-derived metrics to avoid numerical errors. Then, the strain and strain rate calculated based on the kinematics with/without denoising and the reference kinematics were converted to the concussion risk and the error in the estimated concussion risk was analyzed.

**6. Model performance evaluation**

To evaluate the effect of denoising the mouthguard kinematics measurements, we firstly computed the errors on the test sets based on the laboratory impacts collected with the linear pneumatic impactor system. We partitioned the entire dataset of laboratory impacts into the training set, validation set and test set. The entire test set consisted of 25 impacts). Among the 25 impacts, we further selected 9 impacts collected from the second laboratory dataset because these impacts were from unique combinations of impact location and impact velocity. This test subset enabled us to evaluate the model performance on completely unseen impact conditions.

Firstly, on the level of kinematics, we quantified the errors between the the reference and the mouthguard measurements by computing the following metrics: pointwise mean absolute error (MAE), pointwise root mean squared error (RMSE), peak absolute error (PAE), coefficient of determination ($R^2$) of the peak and the signal-to-noise ratio (SNR). The first three pointwise metrics were used to evaluate the ability of the model to generate denoised mouthguard measurements that are closer to the reference measurements for the entire time traces; they were calculated on the values at every time point (100 time points for each 100- ms impact) for each test impact. When reporting the results, we plotted the distribution of pointwise RMSE (the number of data points equals the number of test impacts) in violin plots, the distribution of the errors in Bland Altman plots, and reported the mean MAE, RMSE over the test impacts in a tabular format.

The PAE, the RMSE, $R^2$, Pearson correlation coefficient (PCor) and Spearman correlation coefficient (SCor) of the peaks were used to evaluate the denoising effect on the largest values of the kinematics because the peak values have been frequently used by researchers to summarize the severity of head impacts. These peak-related metrics were calculated on the peak values (one value for each test impact). We plotted the distribution of PAE (the number of data points equals the number of test impacts), and reported the mean PAE, mean peak-RMSE and peak-$R^2$ in a tabular format. Furthermore, the SNR was calculated to directly reflect the effect of denoising on the entire time traces when we assumed the difference between the reference kinematics and the original mouthguard kinematics measurements to be the noise. It is calculated based on the following equations:

$$SNR_{in} = 10 \times log_{10}\left(\frac{\sum_{n=1}^{N} x_i^2}{\sum_{n=1}^{N} (\widetilde{x_i} - x_i)^2}\right)$$

$$SNR_{out} = 10 \times log_{10}(\frac{\sum_{n=1}^{N} x_i^2}{\sum_{n=1}^{N} (\widehat{x_i} - x_i)^2})$$

Here, the $x_i$ denotes the value of the reference kinematics signal at the sampling point i while the x_tilde_i denotes the value of the original kinematics signal at the sampling point i and the $\hat{x}_i$ denotes the value of the denoised kinematics signal at the sampling point i. The $SNR_{in}$ and $SNR_{out}$ denote the SNR for the input signal and output signal, respectively.

On the BIC level, we calculated the absolute error between the BIC or estimated concussion risks calculated by the reference kinematics measurements and the mouthguard kinematics measurements with/without denoising. On the tissue-level strain and strain rate level, the absolute error in the element-wise MPS and MPSR was calculated for each of 4,124 brain elements.

Besides the test sets from the laboratory-collected impacts, we further applied the denoising 1D-CNN models to a previously published on-field football dataset collected with the Stanford instrumented mouthguard [46]. On this blind-test on-field college football impacts, we calculated the difference in the peak magnitude of linear acceleration, the peak magnitude of angular velocity and the whole-brain MPS and MPSR. Furthermore, the distributions of the MPS and MPSR calculated by FEM were also visualized and compared. Data collection was approved by the Institutional Review Boards (IRB, eProtocol number: 45932) at Stanford University.

7. **Exploration of model performance on post-mortem human subject (PMHS) impacts**

In this study, we also performed an exploratory experiment to test whether the 1D-CNN denoising models work in worst-case scenarios with completely unconstrained mandibles. The angular velocity measurements given by the instrumented mouthguards and reference sensors screwed to the back of the skull from 413 PMHS drop test impacts were evaluated. More details of the data collection process for these PMHS impacts can be found in a previous publication [53]. After applying the denoising models on the mouthguard kinematics, the correction after denoising and the noise reduction for the top 10% noisiest impacts and top 5% noisiest impacts in terms of the absolute error of the maximum angular velocity magnitude were evaluated. Accuracy metrics (pointwise RMSE, peak RMSE, peak coefficient of determination, and peak Pearson correlation coefficient) were calculated.

8. **Statistical tests**

In this study, the errors in kinematics, BIC and tissue-level strain and strain rate were compared between the kinematics measurements with/without denoising with 1D-CNN models. Shapiro-Wilk test rejected the normality assumption of the data distribution. Therefore, Wilcoxon signed-rank tests were performed to test the statistical significance of the error distribution difference (with/without denoising) on three levels of model evaluation [47].

## Declaration of Conflict of Interests

The authors declare no conflict of interests in this study.

## Author Contribution

X.Zhan and Y. Liu conceived this study. X. Zhan, Y. Liu, A. Callan, N. Cecchi collected the data. X. Zhan developed the models and performed the data analysis. X. Zhan and Y. Liu wrote the paper manuscript. E. Le Flao, N. Cecchi, O. Gevaert, M. Zeineh, G. Grant, D. Camarillo revised the manuscript. D. Camarillo supervised this study.

## Code availability

The code in the development and deployment of the models reported in this study can be found on the Github repository: https://github.com/xzhan96-stf/mg_denoising

## Acknowledgement

This research was supported by the Pac-12 Conference's Student-Athlete Health and Well-Being Initiative, the National Institutes of Health (R24NS098518), Taube Stanford Children's Concussion Initiative and Stanford Department of Bioengineering. The authors also appreciate the instructions from Dr. Serena Yeung from Stanford CS271: AI for Healthcare.

# Tables

**Table 1. The abbreviations and acronyms used in this study.**

| Abbreviation/Acronym | Meaning |
| --- | --- |
| TBI | mild traumatic brain injury |
| BIC | brain injury criterion |
| HIC | head injury criterion |
| BrIC | brain injury criterion |
| MPS | maximum principal strain |
| MPSR | maximum principal strain rate |
| FEM | finite element modeling |
| HITS | head impact telemetry system |
| ATD | anthropomorphic test device |
| DLHM | deep learning head model |
| 1D-CNN | one-dimensional convolutional neural network |
| Lin. Acc. | Linear Acceleration |
| Ang. Vel. | Angular Velocity |
| MAE | mean absolute error |
| RMSE | root mean squared error |
| PAE | peak absolute error |
| SNR | signal-to-noise ratio |
| MG | mouthguard |
| IRF | injury risk function |
| HIP | head injury power |
| GAMBIT | generalized acceleration model for brain injury threshold |
| SI | severity index |
| MLSH | Mandible Load Sensing Headform |
| PMHS | Post-mortem human subject |

**Table 2. The mean model performance metrics on the kinematics level evaluated on the 25 test impacts.** MAE: mean absolute error (pointwise), RMSE: root mean squared error (pointwise), PAE: peak absolute error, SNR: signal-to-noise ratio, all averaged across test impacts. The metric values indicating the improvement after denoising are bolded.

| **Kinematics** | **Axis** | **Model** | **MAE** | **RMSE** | **PAE** | **Peak-RMSE** | **Peak-$R^2$** | **SNR** | **Peak-r** | **Peak-SCor** |
|---|---|---|---|---|---|---|---|---|---|---|
| Angular Velocity | X | Denoised | 0.71 | 1.32 | 1.08 | 1.71 | 0.99 | 12.33 | 0.99 | 0.98 |
| Angular Velocity | X | Original | 1.55 | 2.54 | 3.71 | 4.38 | 0.90 | 4.94 | 0.99 | 0.98 |
| Angular Velocity | Y | Denoised | 1.17 | 2.05 | 2.01 | 2.60 | 0.90 | 14.57 | 0.96 | 0.82 |
| Angular Velocity | Y | Original | 1.32 | 2.29 | 2.86 | 3.85 | 0.79 | 13.41 | 0.92 | 0.78 |
| Angular Velocity | Z | Denoised | 0.62 | 1.14 | 0.87 | 1.23 | 0.98 | 9.13 | 0.99 | 0.98 |
| Angular Velocity | Z | Original | 0.81 | 1.31 | 1.45 | 1.84 | 0.96 | 7.77 | 0.99 | 0.98 |
| Angular Velocity | Magnitude | Denoised | 0.81 | 1.53 | 1.15 | 1.59 | 0.98 | 22.12 | 0.99 | 0.98 |
| Angular Velocity | Magnitude | Original | 1.06 | 1.76 | 1.98 | 2.37 | 0.94 | 19.06 | 0.99 | 0.98 |
| Linear Acceleration | X | Denoised | 2.34 | 4.53 | 3.51 | 4.63 | 0.92 | 7.97 | 0.97 | 0.92 |
| Linear Acceleration | X | Original | 3.72 | 6.32 | 8.79 | 13.66 | 0.34 | 3.49 | 0.81 | 0.73 |
| Linear Acceleration | Y | Denoised | 1.39 | 2.74 | 2.57 | 3.56 | 0.98 | 5.89 | 0.99 | 0.95 |
| Linear Acceleration | Y | Original | 2.68 | 4.63 | 8.20 | 10.19 | 0.80 | -0.33 | 0.96 | 0.94 |
| Linear Acceleration | Z | Denoised | 1.25 | 2.31 | 3.74 | 5.15 | 0.41 | 7.19 | 0.80 | 0.70 |
| Linear Acceleration | Z | Original | 2.93 | 5.51 | 10.23 | 18.04 | -6.22 | 0.23 | 0.71 | 0.89 |
| Linear Acceleration | Magnitude | Denoised | 1.89 | 3.79 | 3.29 | 4.16 | 0.94 | 13.46 | 0.97 | 0.98 |
| Linear Acceleration | Magnitude | Original | 4.21 | 7.51 | 10.61 | 17.01 | 0.01 | 7.07 | 0.77 | 0.83 |

**Table 3. The mean model performance metrics on the kinematics level evaluated on the 9-impact test subset.** MAE: mean absolute error (pointwise), RMSE: root mean squared error (pointwise), PAE: peak absolute error, SNR: signal-to-noise ratio. The metric values indicating the improvement after denoising are bolded. The 9 test impacts from the o-repeat dataset are particularly selected to evaluate the model performance on completely unseen impact directions and impact velocities.

| **Kinematics** | **Axis** | **Model** | **MAE** | **RMSE** | **PAE** | **Peak-RMSE** | **Peak-$R^2$** | **SNR** | **Peak-r** | **Peak-SCor** |
|---|---|---|---|---|---|---|---|---|---|---|
| Angular Velocity | X | Denoised | 0.97 | 1.725 | 1.975 | 2.637 | 0.926 | 12.641 | 0.97 | 0.817 |
| Angular Velocity | X | Original | 1.49 | 2.63 | 3.87 | 4.96 | 0.74 | 9.10 | 0.96 | 0.95 |
| Angular Velocity | Y | Denoised | 1.47 | 2.53 | 1.99 | 2.34 | 0.88 | 12.32 | 0.96 | 0.93 |
| Angular Velocity | Y | Original | 1.77 | 3.10 | 4.26 | 5.41 | 0.36 | 10.54 | 0.85 | 0.87 |
| Angular Velocity | Z | Denoised | 0.82 | 1.58 | 1.29 | 1.77 | 0.96 | 13.12 | 0.98 | 0.95 |
| Angular Velocity | Z | Original | 1.09 | 1.81 | 2.22 | 2.51 | 0.93 | 11.38 | 0.99 | 0.93 |
| Angular Velocity | Magnitude | Denoised | 1.08 | 2.02 | 1.00 | 1.31 | 0.98 | 20.09 | 0.99 | 0.97 |
| Angular Velocity | Magnitude | Original | 0.97 | 1.85 | 1.94 | 2.79 | 0.91 | 19.74 | 0.98 | 0.92 |
| Linear Acceleration | X | Denoised | 2.58 | 4.53 | 3.52 | 4.14 | 0.78 | 6.62 | 0.93 | 0.87 |
| Linear Acceleration | X | Original | 4.05 | 8.10 | 18.36 | 22.12 | -5.30 | 1.75 | 0.74 | 0.83 |
| Linear Acceleration | Y | Denoised | 1.75 | 3.36 | 3.74 | 3.96 | 0.95 | 7.47 | 0.98 | 0.83 |
| Linear Acceleration | Y | Original | 2.88 | 5.65 | 10.62 | 12.87 | 0.47 | 2.30 | 0.90 | 0.80 |
| Linear Acceleration | Z | Denoised | 1.34 | 2.56 | 5.91 | 7.49 | -2.99 | 7.42 | 0.28 | 0.18 |
| Linear Acceleration | Z | Original | 3.35 | 7.34 | 16.73 | 27.26 | -51.95 | -0.58 | 0.63 | 0.87 |
| Linear Acceleration | Magnitude | Denoised | 2.15 | 3.61 | 3.87 | 4.30 | 0.85 | 12.09 | 0.97 | 0.97 |
| Linear Acceleration | Magnitude | Original | 4.99 | 10.04 | 21.05 | 27.17 | -5.14 | 4.96 | 0.56 | 0.78 |

**Table 4. The different denoising effect across different regions on the strain and strain rate level.** The reduction in the mean error and median error in the estimated MPS and MPSR on the 25 test impacts and 9 test-impact subset. The 9 test impacts from the no-repeat dataset are particularly selected to evaluate the model performance on completely unseen impact directions and impact velocities.

| Brain Region | Reduction in Mean MPS Error | | Reduction in Median MPS Error | | Reduction in Mean MPSR Error | | Reduction in Median MPSR Error | |
|---|---|---|---|---|---|---|---|---|
| | 25 impacts | 9 impacts | 25 impacts | 9 impacts | 25 impacts | 9 impacts | 25 impacts | 9 impacts |
| Brain Stem | 0.004 | 0.012 | <0.001 | 0.002 | 3.519 | 10.868 | 0.295 | 4.122 |
| Corpus Callosum | 0.006 | 0.010 | 0.003 | 0.005 | 0.917 | 3.301 | -0.003 | 0.221 |
| Cerebellum | 0.003 | 0.008 | <0.001 | 0.001 | 2.507 | 7.244 | 0.208 | 2.215 |
| Gray Matter | 0.001 | 0.005 | <0.001 | 0.002 | 1.279 | 4.507 | -0.071 | 0.459 |
| Midbrain | 0.005 | 0.011 | 0.003 | 0.003 | 3.284 | 8.855 | 0.114 | 1.013 |
| Thalamus | 0.003 | 0.005 | 0.002 | 0.003 | 0.614 | 2.012 | 0.198 | 0.639 |
| White Matter | 0.002 | 0.004 | <0.001 | 0.002 | 0.799 | 3.250 | -0.059 | 0.424 |

## Supplementary Materials

**Table S1. The 1D-CNN models for denoising each axis of kinematics measured by the instrumented mouthguard with hyperparameters tuned.**

| Kinematics | Modeling Approach* | Channels in Convolution Stage** | Kernel Size | Initial Learning Rate | Epoch | L2 Regularization |
|---|---|---|---|---|---|---|
| Lin. Acc. X | Direct Modeling | 16,32,64 | 10 | 0.005 | 500 | 0.001 |
| Lin. Acc. Y | Noise Addition | 20,40,80 | 10 | 0.005 | 600 | 0.001 |
| Lin. Acc. Z | Direct Modeling | 32,64,128 | 10 | 0.005 | 300 | 0.001 |
| Ang. Vel. X | Direct Modeling | 16,32,64 | 10 | 0.005 | 500 | 0.001 |
| Ang. Vel. Y | Direct Modeling | 16,32,64 | 10 | 0.005 | 500 | 0.001 |
| Ang. Vel. Z | Noise Addition | 20,40,80 | 10 | 0.01 | 700 | 0.001 |

*: It should be noted that we have tested two different modeling approaches: 1) direct modeling: directly model the denoised kinematics by regarding the reference kinematics measured by the sensors in the ATD as the output; 2) noise addition: model the noise between the original mouthguard kinematics measurements and reference kinematics measurements and add the predicted noise onto the original mouthguard kinematics measurements in the predicting stage.

**: In the design of this study, the number of channels in the transpose convolution stage is just the reverse order of that in the convolution stage.

**Table S2. The comparison among different deep learning models on the entire test set.** The deep neural network (DNN) consists of 5 dense layers, batch normalization layers and dropout layers after the first dense layers with ReLU

activation for the first four layers. Number of hidden units for each layer, the dropout rate and training epoch were tuned in the same manner as for the 1D-CNN. The recurrent neural network with long short-term memory (LSTM) consists of two LSTM layers, two dropout layers, and a dense layer. The number of hidden units, dropout rate, training epochs were tuned on the validation set.

| Kinematics | Component | Model | MAE | RMSE | Peak-MAE | Peak-RMSE | Peak-R2 | SNR | Peak-PCor | Peak-SCor |
|---|---|---|---|---|---|---|---|---|---|---|
| Angular Velocity | X | 1D-CNN | 0.71 | 1.32 | 1.08 | 1.71 | 0.99 | 12.33 | 0.99 | 0.98 |
| Angular Velocity | X | DNN | 5.65 | 8.94 | 8.43 | 11.75 | 0.30 | -1.37 | 0.66 | 0.67 |
| Angular Velocity | X | LSTM | 1.88 | 4.26 | 7.02 | 10.46 | 0.45 | 7.03 | 0.87 | 0.98 |
| Angular Velocity | X | Original | 1.55 | 2.54 | 3.71 | 4.38 | 0.90 | 4.94 | 0.99 | 0.98 |
| Angular Velocity | Y | 1D-CNN | 1.17 | 2.05 | 2.01 | 2.60 | 0.90 | 14.57 | 0.96 | 0.82 |
| Angular Velocity | Y | DNN | 6.16 | 7.58 | 5.78 | 7.27 | 0.23 | 1.10 | 0.69 | 0.67 |
| Angular Velocity | Y | LSTM | 1.83 | 3.47 | 5.79 | 7.76 | 0.12 | 10.16 | 0.76 | 0.60 |
| Angular Velocity | Y | Original | 1.32 | 2.29 | 2.86 | 3.85 | 0.79 | 13.41 | 0.92 | 0.78 |
| Angular Velocity | Z | 1D-CNN | 0.62 | 1.14 | 0.87 | 1.23 | 0.98 | 9.13 | 0.99 | 0.98 |
| Angular Velocity | Z | DNN | 2.24 | 3.64 | 5.02 | 7.58 | 0.28 | 0.45 | 0.83 | 0.83 |
| Angular Velocity | Z | LSTM | 1.04 | 2.31 | 3.22 | 5.74 | 0.59 | 6.24 | 0.89 | 0.97 |
| Angular Velocity | Z | Original | 0.81 | 1.31 | 1.45 | 1.84 | 0.96 | 7.77 | 0.99 | 0.98 |
| Linear Acceleration | X | 1D-CNN | 2.34 | 4.53 | 3.51 | 4.63 | 0.92 | 7.97 | 0.97 | 0.92 |
| Linear Acceleration | X | DNN | 6.06 | 8.43 | 14.65 | 17.85 | -0.13 | 0.06 | 0.69 | 0.48 |
| Linear Acceleration | X | LSTM | 2.35 | 5.07 | 12.64 | 19.12 | -0.30 | 6.67 | 0.59 | 0.78 |
| Linear Acceleration | X | Original | 3.72 | 6.32 | 8.79 | 13.66 | 0.34 | 3.49 | 0.81 | 0.73 |
| Linear Acceleration | Y | 1D-CNN | 1.39 | 2.74 | 2.57 | 3.56 | 0.98 | 5.89 | 0.99 | 0.95 |
| Linear | Y | DNN | 5.27 | 9.04 | 14.78 | 22.08 | 0.05 | -3.29 | 0.60 | 0.66 |
| Acceleration | | | | | | | | | | |
| Linear Acceleration | Y | LSTM | 1.97 | 5.20 | 13.55 | 20.66 | 0.16 | 2.64 | 0.89 | 0.88 |
| Linear Acceleration | Y | Original | 2.68 | 4.63 | 8.20 | 10.19 | 0.80 | -0.33 | 0.96 | 0.94 |
| Linear Acceleration | Z | 1D-CNN | 1.25 | 2.31 | 3.74 | 5.15 | 0.41 | 7.19 | 0.80 | 0.70 |
| Linear Acceleration | Z | DNN | 3.41 | 4.93 | 8.68 | 9.66 | -1.07 | -0.82 | 0.23 | 0.28 |
| Linear Acceleration | Z | LSTM | 1.38 | 2.49 | 4.25 | 6.04 | 0.19 | 6.59 | 0.76 | 0.84 |
| Linear Acceleration | Z | Original | 2.93 | 5.51 | 10.23 | 18.04 | -6.22 | 0.23 | 0.71 | 0.89 |

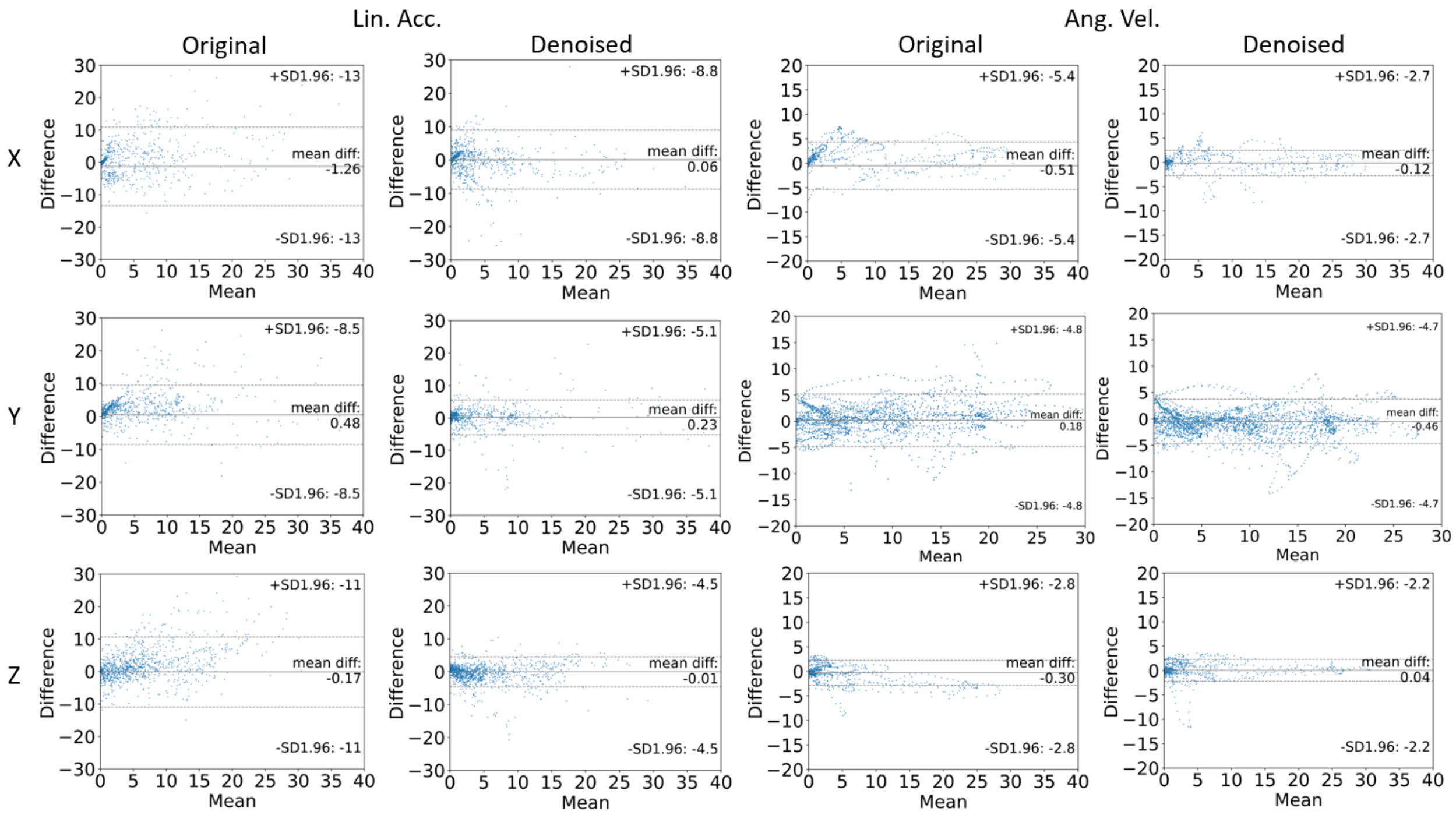

**Figure S1. The Bland Altman plot of the distribution of the pointwise difference between the reference kinematics and mouthguard kinematics with or without denoising on the 25 test impacts.** The difference is calculated in a pointwise manner: 100 time points for each 100ms impact. The x-axis denotes the mean value of the reference kinematics and mouthguard kinematics and the y-axis denotes the deviation (mouthguard kinematics - reference kinematics). The ±1.96SD error bars and the mean deviation are shown with dashed lines.

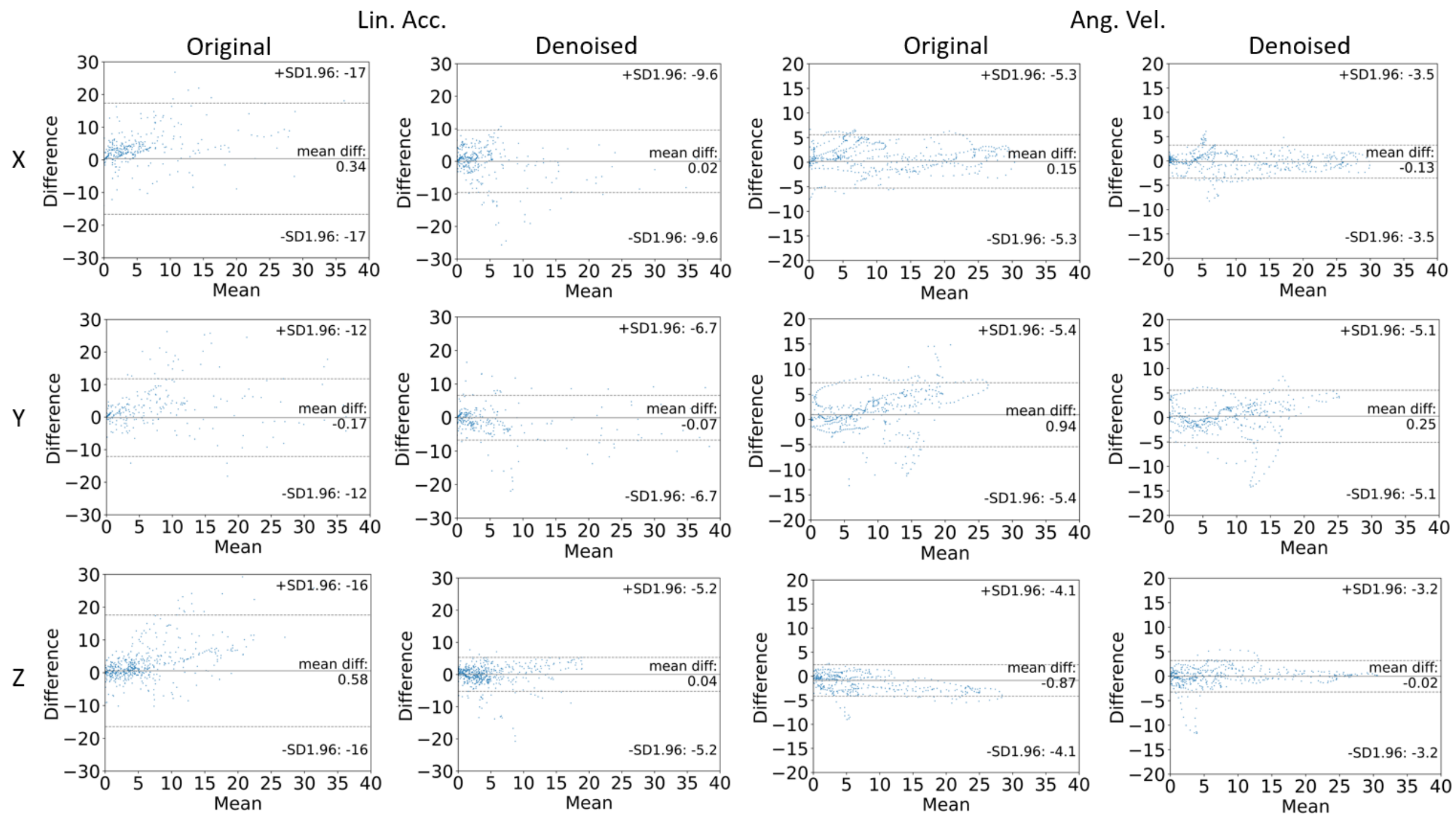

**Figure S2. The Bland Altman plot of the distribution of the pointwise difference between the reference kinematics and mouthguard kinematics with or without denoising on the 9-test-impact subset.** The difference is calculated in a pointwise manner: 100 time points for each 100ms impact. The x-axis denotes the mean value of the reference kinematics and mouthguard kinematics and the y-axis denotes the deviation (mouthguard kinematics - reference kinematics). The ±1.96SD error bars and the mean deviation are shown with dashed lines. The 9 test impacts from the no-repeat dataset are particularly selected to evaluate the model performance on completely unseen impact directions and impact velocities.

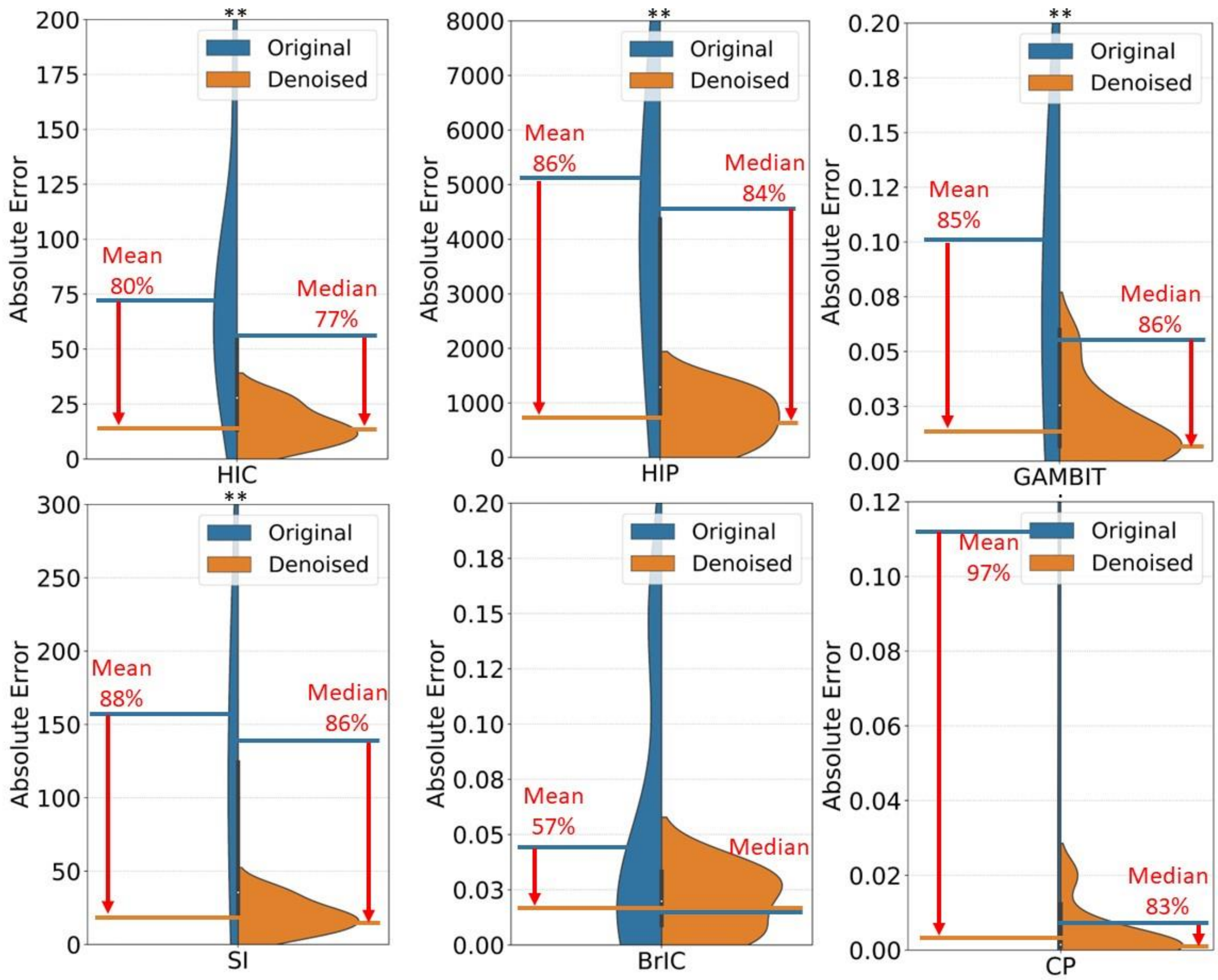

**Figure S3. The distribution of the absolute error between the BIC values calculated by the reference kinematics measurements and that calculated by the mouthguard kinematics measurements with/without denoising on the 9-impact test subset.** The absolute error in six BIC value estimation tasks are shown. The reference BIC values are calculated by the ATD sensor kinematics measurements. The percentage reduction in the error median and the statistical significance between the errors from the original signals and the denoised signals are noted in the plots (.: p<0.1, *:p<0.05, **: p<0.01, ***: p<0.001, Wilcoxon signed-rank test). The 9 test impacts from the no-repeat dataset are particularly selected to evaluate the model performance on completely unseen impact directions and impact velocities.

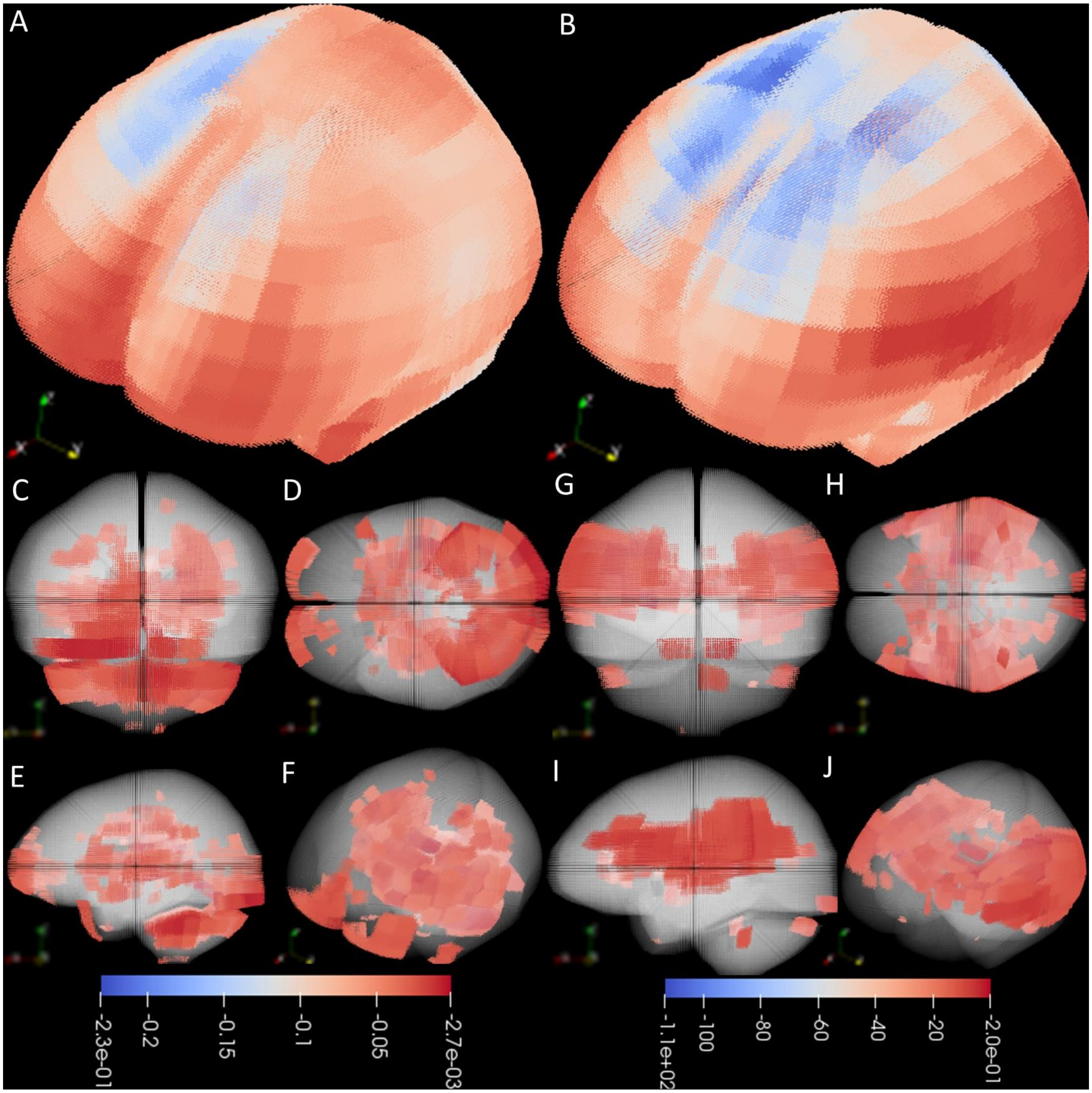

**Figure S4. The three-dimensional visualization of the denoising effect across different brain regions.** The three-dimensional visualization of the correction (values based on the denoised mouthguard kinematics measurements - values based on the original mouthguard kinematics measurements) in the estimation of MPS (left column) and MPSR (right column). The oblique view without transparent brain elements (A, B). The front view (C, G), top view (D, H), side view (E, I) and oblique view (F, J) with the low-value region masked in transparency.

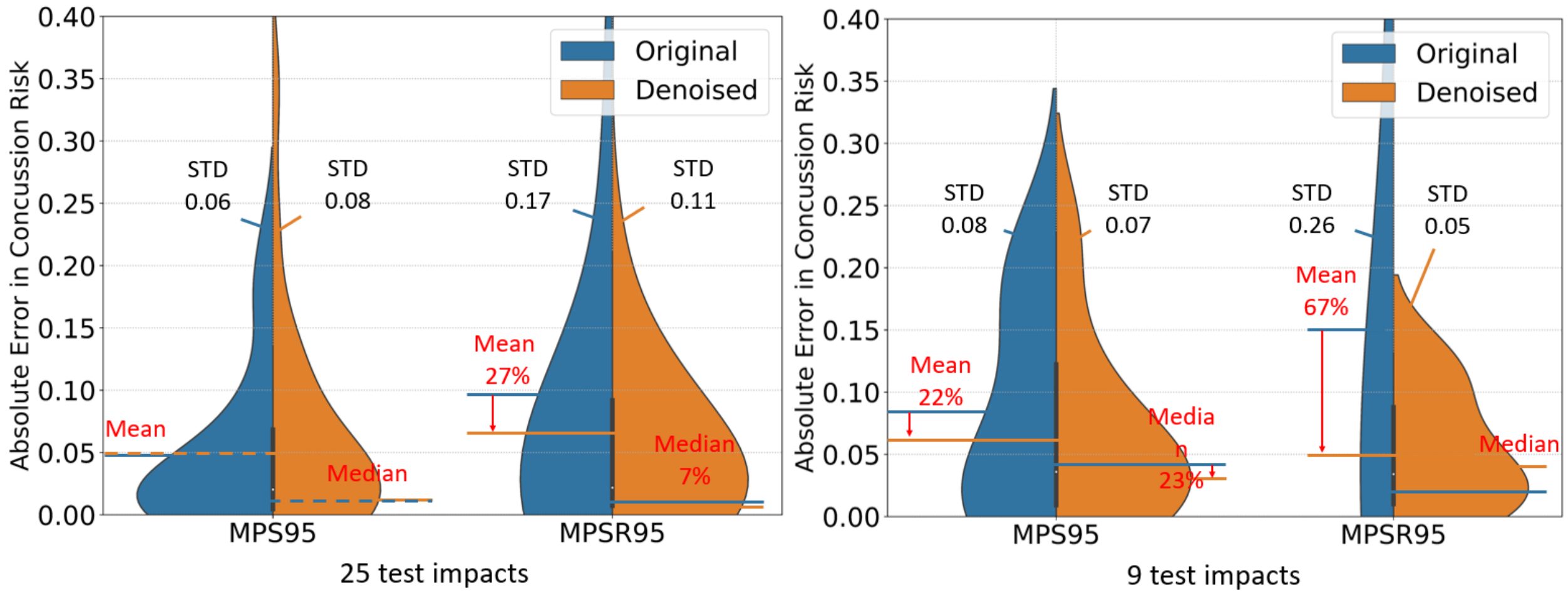

**Figure S5. The distribution of the risk in estimated concussion risk based on the 95th percentile MPS and 95th percentile MPSR.** The risks are calculated by the logistic regression models fitted on the National Football League impact dataset. The relative reduction in mean error and median error and the standard deviation values are marked. The 9 test impacts from the no-repeat dataset are particularly selected to evaluate the model performance on completely unseen impact directions and impact velocities.

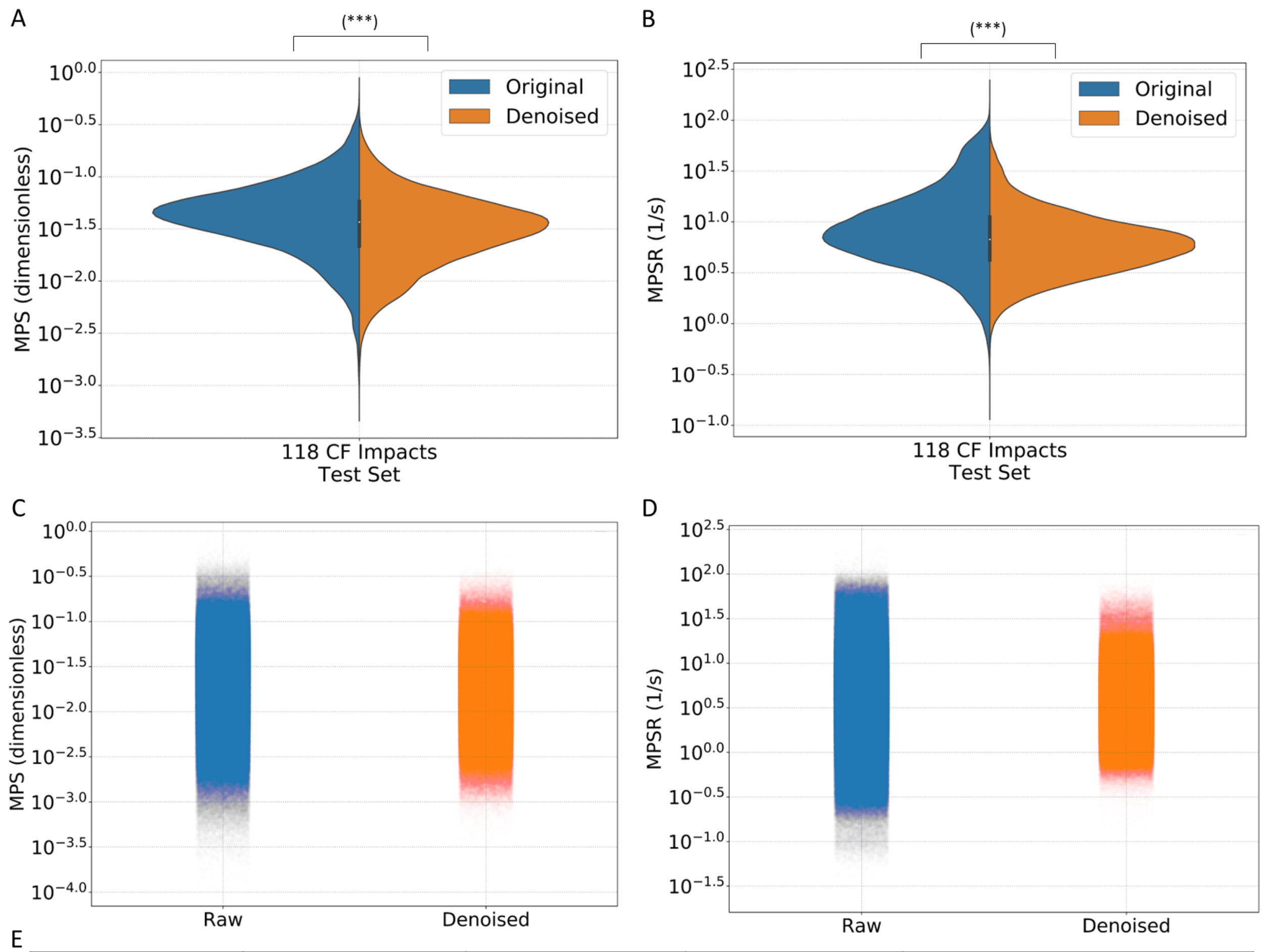

E

| Metric | Method | Median | Mean | Standard Deviation |
|---|---|---|---|---|
| MPS | Original | 0.043 | 0.054 | 0.046 |
| MPS | Denoised | 0.032 | 0.039 | 0.033 |
| MPSR | Original | 8.012 | 12.489 | 13.442 |
| MPSR | Denoised | 5.828 | 7.521 | 6.472 |

**Figure S6. The distribution of the whole-brain element-wise MPS and MPSR estimated based on the mouthguard kinematics with/without denoising on the 118 on-field college football impacts for blind test.** The log-scale violin plots of the element-wise MPS (A) and MPSR (B) and the log-scale strip plots of the element-wise MPS (C) and MPSR (D) with and without denoising with the 1D-CNN models. The mean, median and standard deviation over the distributions are shown in (E). The statistical significance between the errors from the original kinematics signals and the denoised kinematics signals are noted in the plots (.: $p<0.1$, *:$p<0.05$, **: $p<0.01$, ***: $p<0.001$, Wilcoxon signed-rank test).

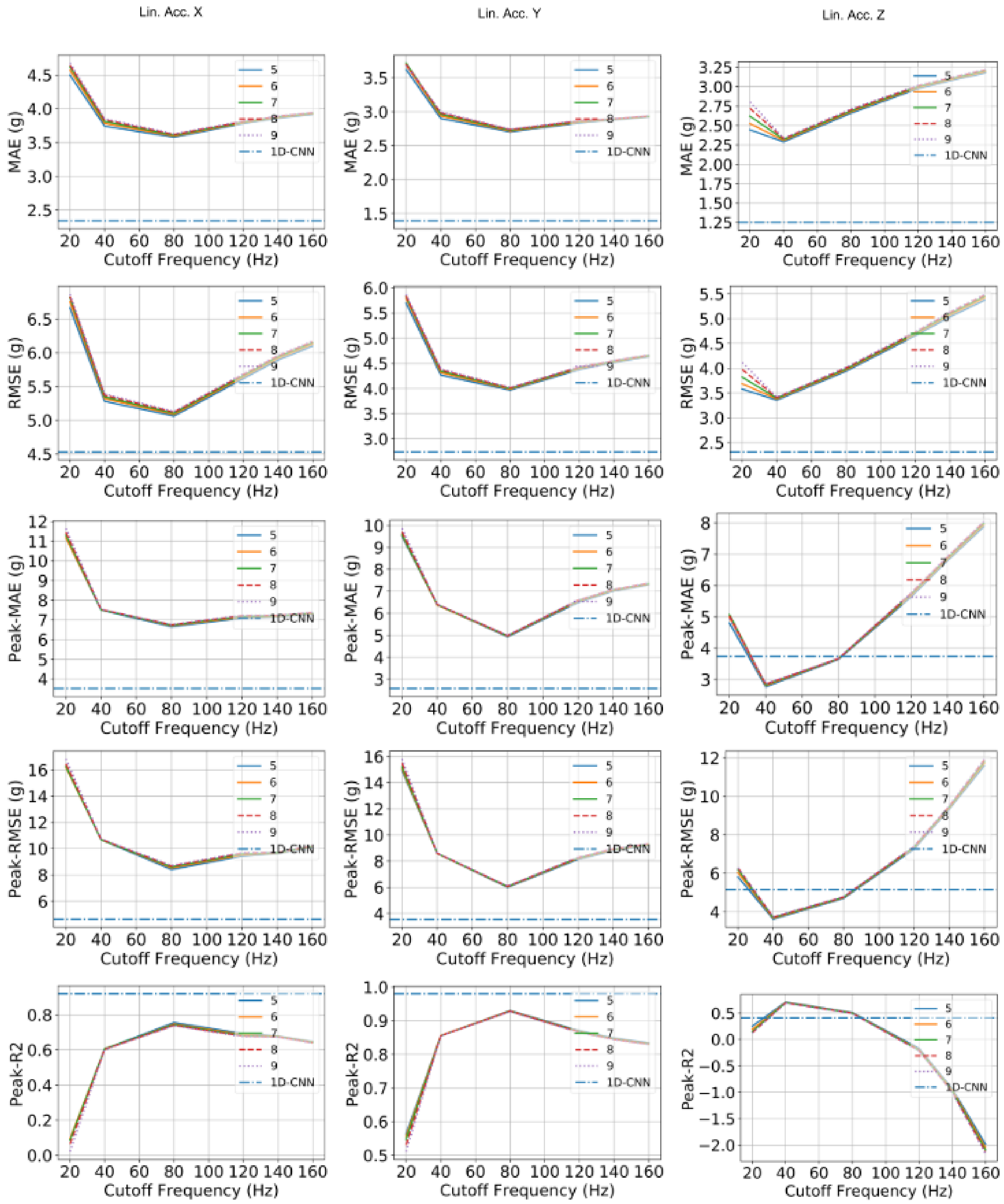

**Figure S7. The comparison of the 1D-CNN models and simple signal filtering approach on denoising the linear acceleration.** Three columns: linear acceleration along the X-axis, Y-axis, Z-axis. The signal filtering was performed with a Butterworth low-pass filter with varying cutoff frequency and order. The metrics were evaluated on the entire test set.MAE: mean absolute error (pointwise), RMSE: root mean squared error (pointwise).

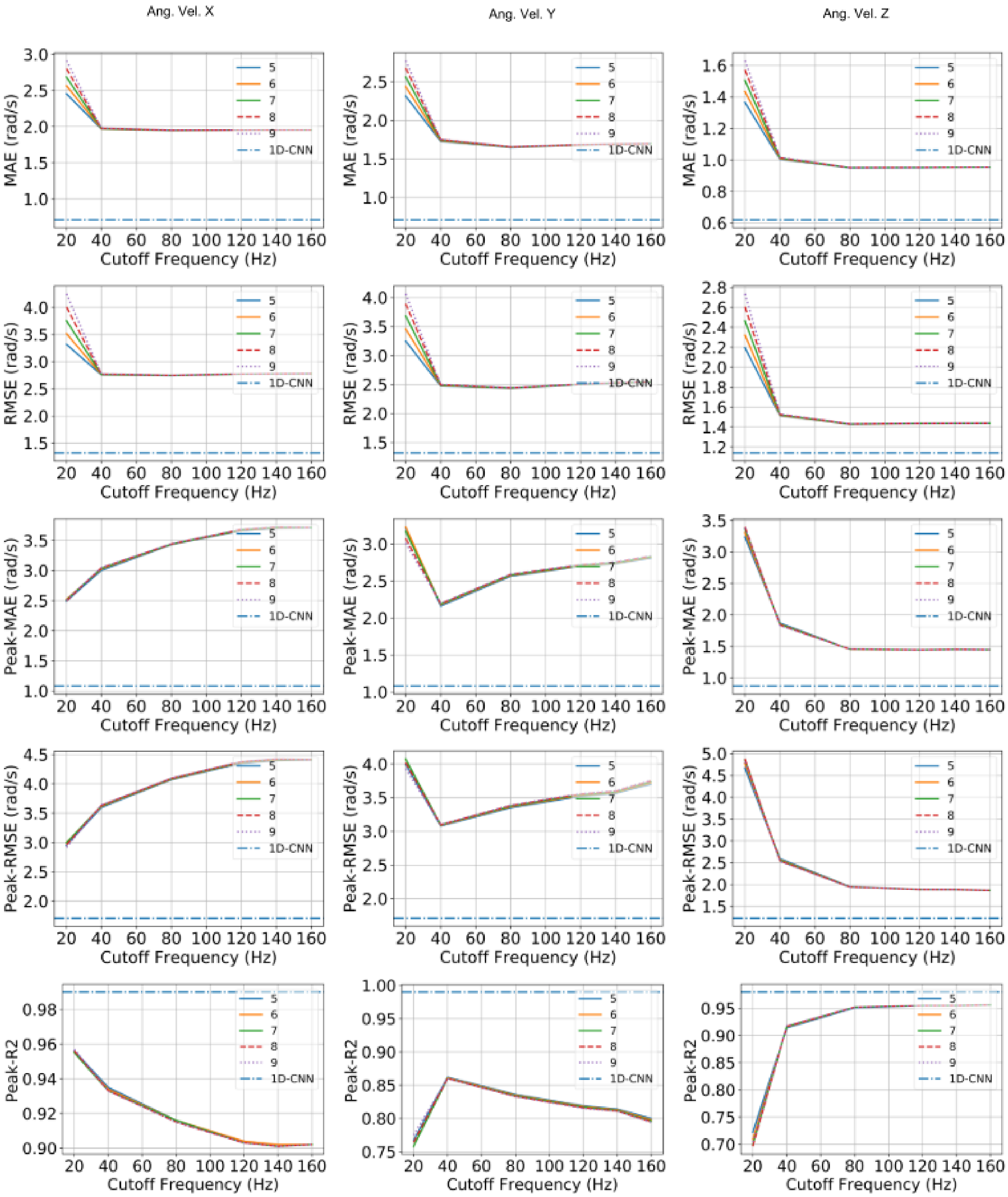

**Figure S8. The comparison of the 1D-CNN models and simple signal filtering approach on denoising the angular velocity.** Three columns: angular velocity along the X-axis, Y-axis, Z-axis. The signal filtering was performed with a Butterworth low-pass filter with varying cutoff frequency and order. The metrics were evaluated on the entire test set.MAE: mean absolute error (pointwise), RMSE: root mean squared error (pointwise).